\documentclass{ecai} 

\usepackage{subfigure}
\usepackage{latexsym}
\usepackage{amssymb}
\usepackage{amsmath}
\usepackage{amsthm}
\usepackage{booktabs}
\usepackage{enumitem}
\usepackage{graphicx}
\usepackage{color}
\usepackage{multirow}

\newcommand{\BibTeX}{B\kern-.05em{\sc i\kern-.025em b}\kern-.08em\TeX}

\let\oldequation\equation
\let\oldendequation\endequation

\renewenvironment{equation}
{\linenomathNonumbers\oldequation}
{\oldendequation\endlinenomath}
\begin{document}

%%%%%%%%%%%%%%%%%%%%%%%%%%%%%%%%%%%%%%%%%%%%%%%%%%%%%%%%%%%%%%%%%%%%%%%%

\begin{frontmatter}

%%% Use this command to specify your submission number.
%%% In doubleblind mode, it will be printed on the first page.

\paperid{3678} 

%%% Use this command to specify the title of your paper.
%%% Use this combinations of commands to specify all authors of your 
%%% paper. Use \fnms{} and \snm{} to indicate everyone's first names 
%%% and surname. This will help the publisher with indexing the 
%%% proceedings. Please use a reasonable approximation in case your 
%%% name does not neatly split into "first names" and "surname".
%%% Specifying your ORCID digital identifier is optional. 
%%% Use the \thanks{} command to indicate one or more corresponding 
%%% authors and their email address(es). If so desired, you can specify
%%% author contributions using the \footnote{} command.
	
\title{ScSAM: Debiasing Morphology and Distributional Variability in Subcellular Semantic Segmentation}

\author[A]{\fnms{Bo}~\snm{Fang}}
\author[A]{\fnms{Jianan}~\snm{Fan}}
\author[A]{\fnms{Dongnan}~\snm{Liu}} % use of \orcid{} is optional
\author[B]{\fnms{Hang}~\snm{Chang}}
\author[C,D]{\fnms{Gerald J.}~\snm{Shami}}
\author[C,D]{\fnms{Filip}~\snm{Braet}}
\author[A]{\fnms{Weidong}~\snm{Cai}}

\address[A]{School of Computer Science, University of Sydney, Australia}
\address[B]{Lawrence Berkeley National Laboratory, USA}
\address[C]{School of Medical Sciences (Molecular and Cellular Biomedicine), University of Sydney, Australia}
\address[D]{Australian Centre for Microscopy and Microanalysis, University of Sydney, Australia}

%%% Use this environment to include an abstract of your paper.

\begin{abstract}
The significant morphological and distributional variability among subcellular components poses a long-standing challenge for learning-based organelle segmentation models, significantly increasing the risk of biased feature learning.
Existing methods often rely on single mapping relationships, overlooking feature diversity and thereby inducing biased training.
Although the Segment Anything Model (SAM) provides rich feature representations, its application to subcellular scenarios is hindered by two key challenges: (1) The variability in subcellular morphology and distribution creates gaps in the label space, leading the model to learn spurious or biased features.
(2) SAM focuses on global contextual understanding and often ignores fine-grained spatial details, making it challenging to capture subtle structural alterations and cope with skewed data distributions.
To address these challenges, we introduce ScSAM, a method that enhances feature robustness by fusing pre-trained SAM with Masked Autoencoder (MAE)-guided cellular prior knowledge to alleviate training bias from data imbalance.
Specifically, we design a feature alignment and fusion module to align pre-trained embeddings to the same feature space and efficiently combine different representations.
Moreover, we present a cosine similarity matrix-based class prompt encoder to activate class-specific features to recognize subcellular categories.
Extensive experiments on diverse subcellular image datasets demonstrate that ScSAM outperforms state-of-the-art methods.
\end{abstract}

\end{frontmatter}

%%%%%%%%%%%%%%%%%%%%%%%%%%%%%%%%%%%%%%%%%%%%%%%%%%%%%%%%%%%%%%%%%%%%%%%%
\begin{figure}[t]
	\centering
	\subfigure[Traditional Feature Fusion Segmentation Model]{
		\includegraphics[width=0.45\textwidth]{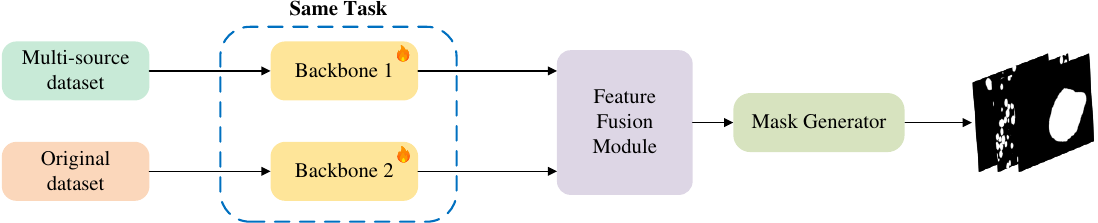}
		\label{fig:tffsm}
	}
	\subfigure[SAM-based Feature Fusion Segmentation Model]{
		\includegraphics[width=0.45\textwidth]{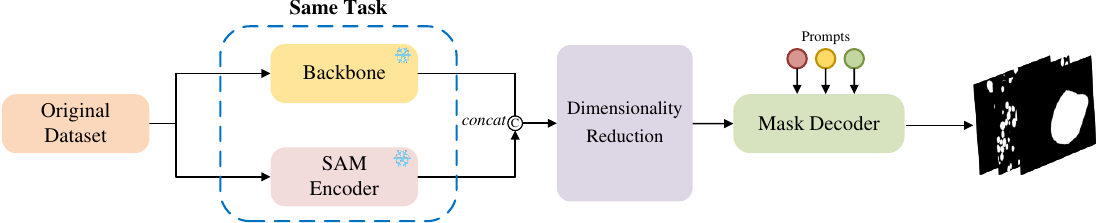}
		\label{fig:sffsm}
	}
	\subfigure[ScSAM]{
		\includegraphics[width=0.45\textwidth]{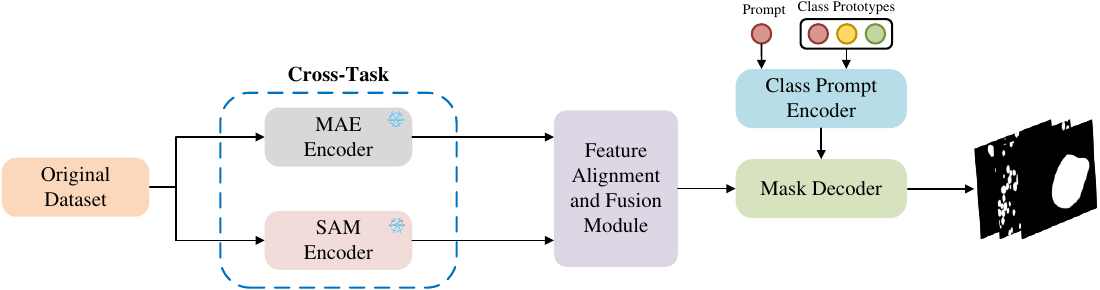}
		\label{fig:scsam}
	}
	\caption{Comparison of ScSAM against existing feature fusion segmentation models. Traditional methods need to update the parameters of backbones and require additional relevant datasets, whereas SAM-based fusion algorithms contain frozen backbones pre-trained for the same task. The proposed ScSAM fuses cross-task feature representations to problematic learning patterns in subcellular segmentation.}
	\label{fig:combined}
\end{figure}
\section{Introduction}
Electron microscopy reveals the intricate nanoscale universe within living cells, capturing the morphology and distribution of organelles from microscopic particles to massive nuclei.
Precise subcellular segmentation is pivotal for cell behavior studies, disease mechanism resolution, and drug development, in addition to examining intra- and inter-cellular interactions \cite{xie2024domain,sekh2021nature}.
Nevertheless, due to the diverse morphology and extreme spatial heterogeneity of subcellular structures, conventional subcellular recognition techniques fail to depict accurate contours \cite{chan2001active,boykov2006graph}.
Consequently, methods designed to debias morphology and distributional variability are urgently required for microscopic image analysis \cite{rizk2014nature,bissoto2020debiasing}.

The Segment Anything Model (SAM) \cite{kirillov2023iccv}, as a prominent large-scale foundation model, has attracted widespread attention for its excellent promptable segmentation capabilities.
Adopting SAM for various application scenarios, such as segmentation, detection, and tracking, can improve training efficiency while leveraging robust pre-training knowledge \cite{zhou2024aaai,abdur2024sam}.
Nevertheless, due to the substantial gap between natural and target domains, applying SAM to segmentation tasks in a zero-shot manner fails to generate satisfactory predictions.
Due to the limitations of statistical distributions and structural priors, many previous works reported that SAM performs poorly in zero-shot segmentation applications such as surgical instruments, medical images, and optical flow analysis \cite{yue2024aaai,zhou2024aaai}.

Specifically, applying SAM directly to microscopic image analysis faces two key challenges: 1) Morphological and distributional variability leads to data imbalance in the label space, inducing biased parameter optimization, especially for organelles such as granules, which have small-scale contours and irregular distributions; and
2) SAM focuses on global contextual understanding and ignores local spatial details.
This single mapping relation cannot exhibit its full potential when applied to specific scenarios, making it difficult to handle intricate subcellular morphologies and subtle features \cite{cheng2024unleashing,zhang2024distilling}.

Recent endeavors to tackle the model transfer challenge aim to apply the general-purpose model to specific domains by expanding feature diversity and fusing representations from multiple backbones.
As illustrated in Fig.~\ref{fig:combined}, traditional frameworks enrich micro-texture and macro-semantics to improve the precision of the final predictions in a homogeneous ensemble manner \cite{li2023cvpr,wang2019pseudo}.
However, the numerous training parameters of the traditional framework reduce efficiency and increase computational burden. 
Although SAM-based feature fusion segmentation models reduce the training parameters, same-task backbones learn highly redundant feature representations with limited semantic diversity, limiting their ability to understand and characterize complex scenes, especially when dealing with highly overlapping and irregularly distributed subcellular structures \cite{zhang2023miccai,zhou2024aaai}.

\begin{figure}[t]
	\centering
	\hspace*{\fill}
	\hspace*{\fill}
	\subfigure[Masked frame]{
		\includegraphics[width=0.135\textwidth]{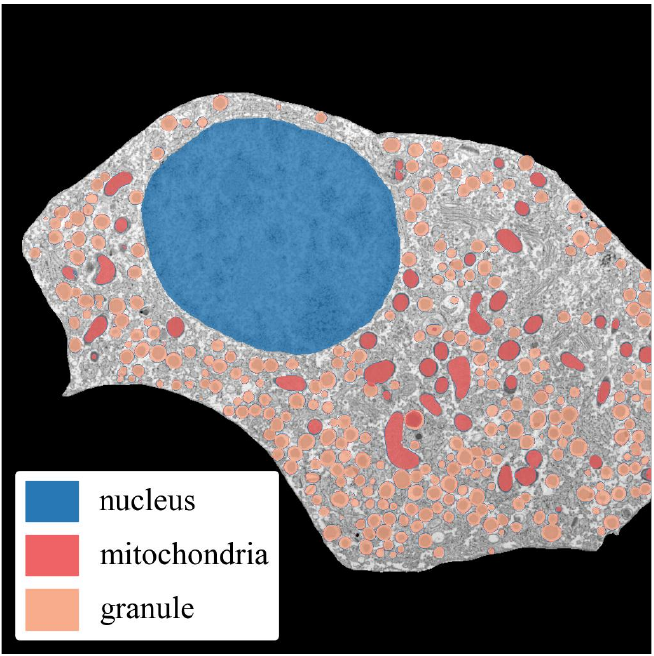}
		\label{fig:r1}
	}
	\hfill
	\subfigure[SAM]{
		\includegraphics[width=0.135\textwidth]{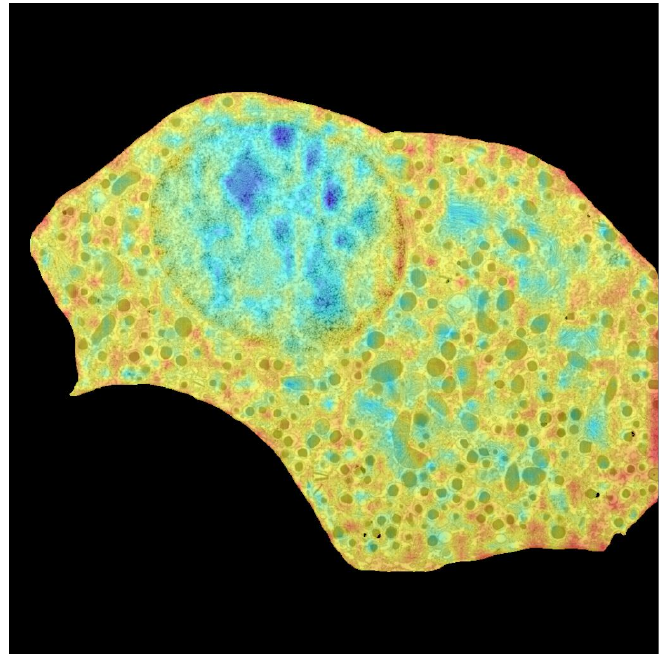}
		\label{fig:r2}
	}
	\hfill
	\subfigure[MAE]{
		\includegraphics[width=0.135\textwidth]{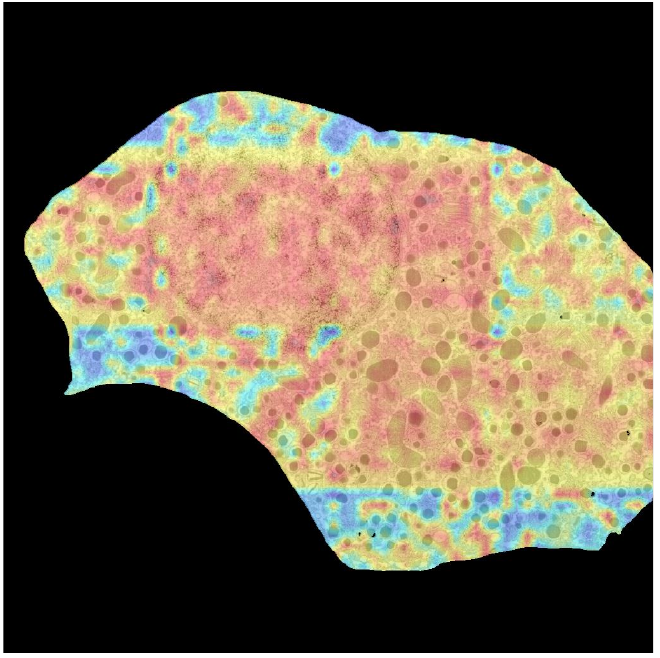}
		\label{fig:r3}
	}
	\hspace*{\fill}
	\hspace*{\fill}
	\caption{Visualization of subcellular distribution and feature activation maps: (a) Masked frame, (b) and (c) are feature activation maps for SAM and MAE embeddings indicating complementary feature representations. The red response region is the region of interest of dual backbones, demonstrating orthogonal information dimensions in the embedders.}
	\label{fig:prompting}
\end{figure}

To address the aforementioned issues, we introduce ScSAM, an end-to-end subcellular segmentation framework designed to handle complex data distribution scenarios (as shown in Fig.~\ref{fig:r1}).
Technically, we devise a dual structure with encoders trained on distinct tasks to fuse complementary semantic cues and faithfully capture the pronounced morphological and spatial heterogeneity of subcellular organelles.
Figs.~\ref{fig:r2} and \ref{fig:r3} display the intra-cellular activation maps from two encoders, revealing different and complementary feature representations.
Specifically, the Masked Autoencoder (MAE) attends to multi-scale structural patterns, spanning subtle local textures and intermediate morphological motifs to overarching global arrangements, while SAM aims to extract structure-related features such as edges, shapes, and region-level consistency.
For semantic spatial synergy, we propose a Feature Alignment and Fusion Module (FAFM) to align and fuse cross-task feature embeddings from two encoders and recalibrate their spatial contributions via attention-driven weighting to enhance fine-grained feature representation (as illustrated in Fig.~\ref{fig:scsam}).
FAFM employs a cosine-similarity loss to align spatial feature directions and alleviate cross-task semantic bias, while a Channel Attention Module (CAM) adaptively re-weights channels to accentuate discriminative cues.
To eliminate explicit prompts, we devise a class prompt encoder with a residual structure to activate class-aware features by comparing the similarity between learnable class prototypes with visual embeddings.

Our main contributions are summarized as follows:
\begin{itemize}
	\item We develop a novel framework for subcellular recognition in electron microscopy scenarios that, for the first time, fuses cross-task feature representations to enhance its ability to understand and characterize overlapping and irregular subcellular structures.
	
	\item We design the Feature Alignment and Fusion Module (FAFM) that aligns SAM and MAE embeddings to the same feature space and fuses them to integrate local spatial information and high-level semantic features.
	
	\item  We propose a residual class prompt encoder that compares learnable class prototypes with visual embeddings, activating class-aware regions and providing dense and sparse category embeddings for precise organelle discrimination.
	
	\item  We comprehensively evaluate ScSAM in the high- and low-glucose BetaSeg datasets, achieving state-of-the-art (SOTA) performance with limited labeled EMIs.
	
\end{itemize}

\section{Related Work}
\begin{figure*}[t]
	\centering
	\includegraphics[width=0.96\textwidth]{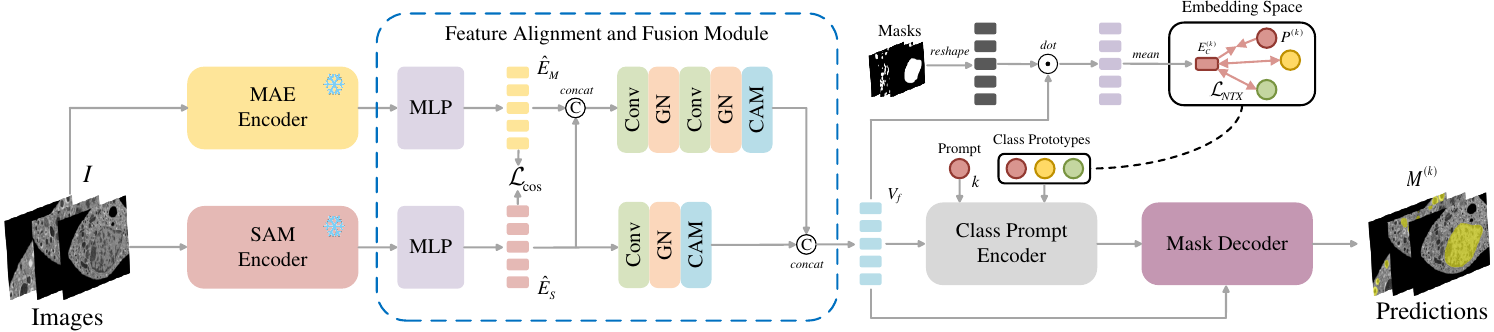}
	\caption{The overview of ScSAM, which utilizes pre-trained MAE as prior knowledge related to subcellular for frozen SAM to enhance feature perception of organelle morphology and distribution in EMIs. We design the FAFM and cosine similarity-based class prompt encoder to deeply fuse and learn organelle class-specific features.}	
	\label{fig:ScSAM} 
\end{figure*}
\subsection{Subcellular Segmentation}
Subcellular segmentation serves as an essential analysis tool in disease research, drug discovery, and biological cellular analysis.
Traditional subcellular segmentation methods assume that intensity gradients coincide with object boundaries, using unsupervised binarization methods such as minimum error thresholding or Otsu's single-level method \cite{otsu1975automatica} to depict subcellular contours from EMIs.
However, these threshold-based traditional methods often fail to capture subtle morphological variations in complex microscopy scenes, perform poorly on overlapping structures, and in low-contrast regions.

Recent advances in feature extraction and learning algorithms have laid a solid foundation for computational approaches in biological image analysis \cite{fan2023taxonomy,fan2024seeing,peng2024sam}.
For instance, TransNuSeg \cite{he2023miccai} and FragViT \cite{luo2024aaai} employ Transformers to capture global context and long-range dependencies for segmenting nuclei and mitochondria in EMIs.
Nevertheless, focusing on individual subcellular structures in isolation cannot support comprehensive behavioral analysis among various cellular components, which is not conducive to unfolding downstream tasks such as pathological state assessment and cellular functional analysis.
A self-supervised voxel-level representation learning method \cite{han2022cvpr} is designed to decompose the latent space into semantic and transformational subspaces, utilizing the representations for unsupervised segmentation of several organelles.
Similarly, a self-supervised method called MAESTER \cite{xie2023cvpr} is proposed to recognize subcellular structures using K-means clustering of MAE embeddings and employs a pixel-by-pixel inference phase to generate prediction masks.
However, these methods either focus exclusively on a single organelle, which hinders holistic cellular analysis, or they perform poorly on smaller organelles, depicting imprecise contours that impede comprehensive interpretation.

\subsection{Segment Anything Model for Customized Tasks}
Recently, SAM \cite{kirillov2023iccv} has gained considerable attention as a vision foundation model, exhibiting excellent zero-shot generalization ability after training on large-scale datasets \cite{lin2024cvpr,yamagiwa2024wacv}.
It can be effectively adapted to different scenarios by providing suitable prompts.
However, providing accurate explicit points or bounding boxes to SAM is challenging, and manual or detector-generated prompts cannot bridge the prior knowledge gap inherent in migration scenarios. 

To address the above issues, some researchers fuse domain-specific information into SAM by incorporating suitable adapters \cite{zhang2024cvpr,chen2023iccv}.
Other researchers design semantic augmentation and multi-layer feature fusion modules to tackle detail-aware customized tasks \cite{yuan2024cvpr,zhang2024cvpr}.
Nevertheless, these methods focus on employing complex structures to extract SAM encoder-based embeddings while ignoring feature information in other dimensions.
Based on this, SAMFlow \cite{zhou2024aaai} and SAM-Path \cite{zhang2023miccai} are proposed to embed other backbones designed for the same task into SAM to enhance object perception, enriching task-specific semantic features and effectively alleviating the issue of visual pattern gap on target recognition.
While these SAM-based algorithms acquire prior knowledge for specific scenes, they neglect the potential auxiliary role of other feature spaces in semantic segmentation tasks.
As for subcellular segmentation, we find no previous research involving fine-tuning SAM for electron microscopy applications.
Therefore, in this paper, we will deeply conduct a thorough investigation of the specific SAM-based framework for EMIs.

\section{Methodology}
\subsection{ScSAM Overview}
Figure~\ref{fig:ScSAM} outlines our proposed ScSAM, where 'Sc' denotes subcellular.
Our primary goal is to exploit multi-scale and complementary cues to enrich the semantic representations, enabling the model to cope with the pronounced morphological diversity and uneven spatial distribution of subcellular structures.
In a nutshell, for any input image $I$ and an organelle class $k$, ScSAM generates a class-specific prediction mask ${M}^{(k)}$, which is defined as:
\begin{equation}
	\label{eq: ove}
	{M}^{(c)}= ScSAM(I, k).
\end{equation}

As indicated in Fig.~\ref{fig:ScSAM}, ScSAM contains three main components: two frozen encoders, a Feature Alignment and Fusion Module (FAFM), and a cosine similarity-based class prompt encoder.
These components are elaborated in the following subsections.

\subsection{Feature Alignment and Fusion Module}
Subcells exhibit significant variability in their morphological features and spatial distribution, posing challenges to the learning and generalization capabilities of segmentation algorithms \cite{fan2024revisiting, song2024cell, liu2020pdam}.
As shown in Fig.~\ref{fig:prompting}, two embeddings are highly heterogeneous in terms of statistical distribution, semantic granularity, and attention patterns due to different pre-training goals.
To harmonize scales and match semantics, we propose the Feature Alignment and Fusion Module (FAFM) to align and fuse embeddings with different feature representations, alleviating the biased learning issue resulting from subcellular imbalance.

As shown in Fig.~\ref{fig:ScSAM}, we first employ two-layer MLPs with an output channel of 256 to align the embeddings to the same dimension:
\begin{equation}
	\label{eq: MLP}
	\hat{E}_{S}= L_1(ReLU(L_2(E_S))),
\end{equation}	
where $\hat{E}_{S}$ and $E_S$ are aligned and original SAM embedding, while $L_1$ and $L_2$ represent the linear projection functions.
As in Eq.~\ref{eq: MLP}, the MAE embedding $E_M$ reduces the dimension to yield $\hat{E}_{M}$, matching $\hat{E}_{S} \in \mathbb{R}^{H \times W \times N}$, where $H \times W$ represents the spatial resolution and $N$ is the number of channels.
We employ the cosine similarity loss to quantify the variation between two resized embedding representations, which is expressed as:
\begin{equation}
	\label{eq: l_cos}
	\mathcal{L}_{cos}= 1 - \frac{1}{N} \sum_{i=1}^{N} \left( \frac{\hat{E}_{S_i} \cdot \hat{E}_{M_i}}{\|\hat{E}_{S_i}\| \|\hat{E}_{M_i}\|} \right),
\end{equation}	
where ${\|\hat{E}_{S_i}\|}$ and ${\|\hat{E}_{M_i}\|}$ are the L2 normalization of the $i$-th embedding vectors of  $\hat{E}_{S}$ and $\hat{E}_{M}$.
ScSAM minimizes $\mathcal{L}_{cos}$ to align the directions of cross-task embeddings while ignoring magnitudes, projecting them into a common submanifold and preserving the distributional spread of each feature space.

Then, a fusion module containing two branches is designed to implement cross-task semantic compensation and downstream adaptation for cosine-aligned embeddings, as shown in Fig.~\ref{fig:ScSAM}.
It first concatenates $\hat{E}_{S}$ and $\hat{E}_{A}$ along the channel dimension and feeds the result into two 3$\times$3 convolutional layers and Group Normalization.
Since electron microscopy batches are typically small, GroupNorm avoids the statistical instability exhibited by BatchNorm in this case and maintains the consistency of the feature distribution.
Then the convolutional output enters the channel attention module (CAM) \cite{woo2018cbam} that employs global average and max pooling to collect channel statistics and map them to learnable weights that dynamically emphasize channels associated with organelles.
The vector $V_{c} \in \mathbb{R}^{H \times W \times N/2}$ is obtained from this branched line operation, which is calculated as:
\begin{equation}
	V_c = \text{CAM}(\text{Conv}(\text{Conv}(\text{Concat}(\hat{E}_{A}, \hat{E}_{S})))).
\end{equation}
Since SAM embeddings contain rich subcellular semantic representations, FAFM adds a lightweight auxiliary stream to downsample $\hat{E}_{A}$ and concatenate it with $V_{c}$ as a downstream feature vector to get $V_{f}\in \mathbb{R}^{H \times W \times N}$:
\begin{equation}
	V_f = \text{Concat}(V_c, \text{CAM}(\text{Conv}(\hat{E}_{A}))).
\end{equation}
Overall, FAFM enhances downstream subcellular segmentation by integrating the heterogeneous embeddings into a common feature space and fusing boundary-aware semantics and a priori texture knowledge into a dense representation.
\subsection{Cosine Similarity based Class Prompt Encoder}
\begin{figure}[t]
	\centering 
	\includegraphics[width=0.45\textwidth]{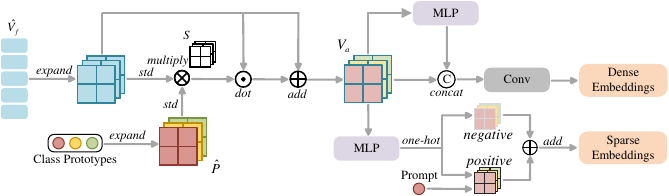}
	\caption{Overview of the class prompt encoder. We present a framework without manual prompts that constructs a cosine similarity matrix to activate class features and generate dense embeddings with a residual connection structure.}
	\label{fig:cpe}
\end{figure}

To eliminate the need for manual prompts, we introduce a class prompt encoder that generates learned prompts for each class.
This module produces a sparse embedding (analogous to a point prompt) and a dense embedding (analogous to a mask prior) for each class, based on the fused feature map.
Inspired by \cite{yue2024aaai}, we introduce a trainable class prototype bank that can hold category information via embedding layer parameter updates.
As shown in Fig.~\ref{fig:cpe}, the cosine similarity between the fused vectors and the class prototype embeddings is computed through matrix multiplication, projecting the fused vectors into the class space associated with each specific organelle.
Then, we enhance the activated feature vector with an MLP-based residual connection for category adaptation and construct class-based positive and negative samples using one-hot coding to generate the dense and sparse embedding demanded by the mask decoder.
Sparse embeddings provide high-confidence local anchors that contain category-aware information based on prompts, while dense embeddings contain the shape and texture knowledge to drive the decoder to refine the boundaries.
This framework effectively implements adaptive class embedding learning, providing abundant semantic information to the decoder and enhancing the feature representation of complex spaces.

Specifically, the class prototype bank is denoted as $P = \{P^{(1)}, P^{(2)},..., P^{(c)}\}$, where $c$ represents the number of classes and $P^{(k)}\in \mathbb{R}^{N}$ is the prototype embedding of class $k$.
The cosine similarity matrix $S^{(k)}$ is computed as the dot product between vectors after L2 normalization.
We measure the similarity between the fused feature $ \hat{V}_f$ and each class prompt $\hat{P}^{(k)}$ using cosine similarity:
\begin{equation}        
	S^{(k)} =\hat{V}_f \times \hat{P}^{(k)}, \quad \text{for } k \in \{1, 2, \ldots, c\},
\end{equation}
where both vectors are $\ell_2$-normalized beforehand.
Then, we modulate the fused feature $\hat{V}_f$ with each class similarity score $S^{(k)}$ to generate class-specific activated features.
The class-specific features $\hat{V}_a$ are activated using the similarity matrix $S^{(k)}$ multiplied with $\hat{V}_f$ and summed:
\begin{equation}        
	{V}_a =\hat{V}_f \cdot S^{(k)}+\hat{V}_f, \quad \text{for } k \in \{1, 2, \ldots, c\}.
\end{equation}
The adjusted feature $V_a$ is forwarded to the segmentation head to generate the final masks.
Then, we design an MLP-based residual connection structure and construct pairs of positive and negative samples by one-hot coding to obtain dense and sparse embeddings that match the mask decoder input shape.

In addition, we analyze class prototypes (anchors) and class embeddings (samples) based on fused vectors to improve intra-class consistency and enhance inter-class separation by employing NTXentLoss, which is defined as:
\begin{equation}        
	\mathcal{L}_{NTX} = - \log \frac{\exp(\text{sim}(P^{(k)}, E _C^{(k)}) / \tau)}{\sum_{j=1}^{2B} 1_{[k \neq i]} \exp(\text{sim}(P^{(k)}, E _C^{(j)})/ \tau)},
\end{equation}
where $\tau$ is the temperature parameter for scaling similarity, while $P^{(k)}$ and $E _C^{(k)}$ are the prototype and class embedding of class $k$.
As shown in Fig.~\ref{fig:ScSAM}, $\mathcal{L}_{NTX}$ effectively clusters samples of the target class in the embedding space while mitigating the similarity between samples of different classes during parameter optimization, thereby enhancing the structural properties of the feature space.
The loss function of ScSAM contains three items: cosine similarity loss for aligned process, NTXentLoss for prototype learning, and Dice loss for semantic segmentation.
It is defined as:
\begin{equation}        
	\mathcal{L} = \lambda\mathcal{L}_{cos} + \mathcal{L}_{NTX} + \mathcal{L}_{Dice},
\end{equation}
where $\lambda$ is a weighting factor of ${L}_{cos}$.
Since ${L}_{cos}$ starts with a gradient decrease from 1, we set its coefficient to 0.2 to alleviate its effect on ScSAM parameter updates in the early optimization.

\section{Experiments and Results}
\begin{figure*}[t]
	\centering
	\includegraphics[width=0.95\textwidth]{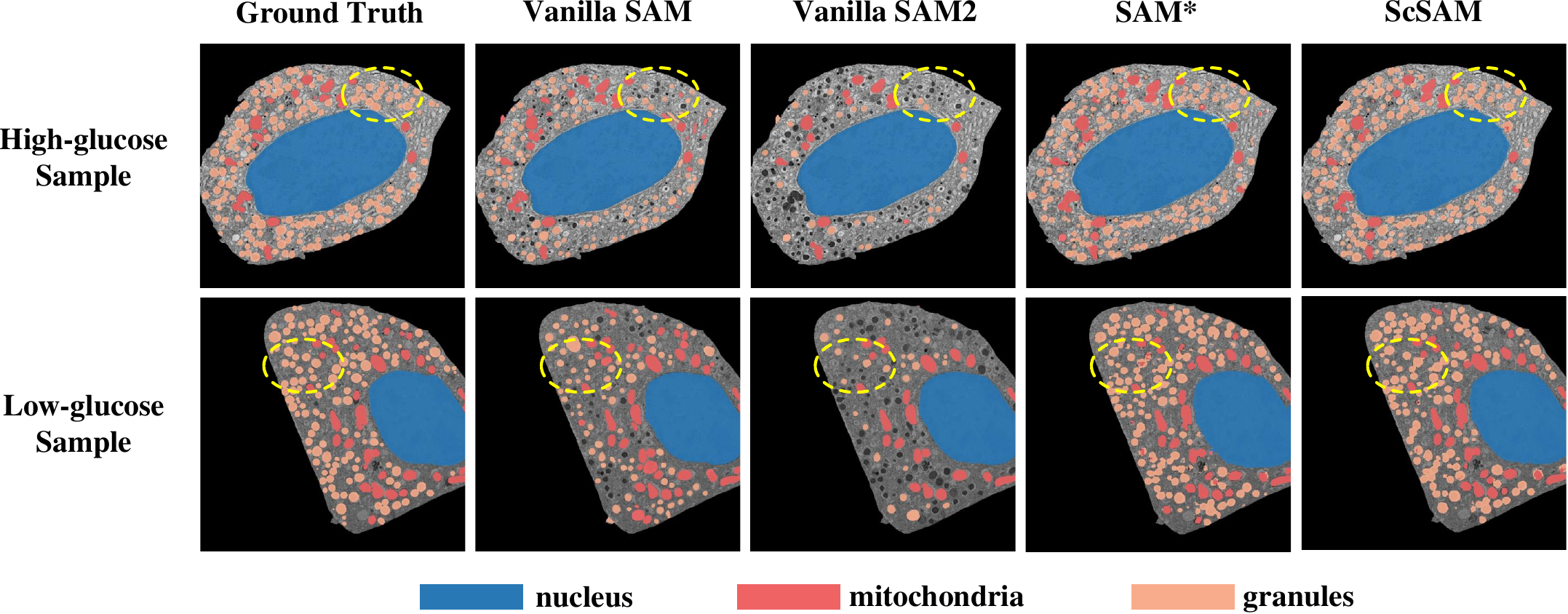}
	\caption{Visual comparison of prediction masks on high- and low-glucose samples. These methods validated one islet cell in each dataset by overlaying the original images and segmentation masks. Notably, the yellow dashed ellipse is used to emphasize regions with significant recognition variance.}	
	\label{fig:results} 
\end{figure*}
\begin{figure}[t] % h表示图片放在当前位置
	\centering % 图片居中
	\includegraphics[width=0.48\textwidth]{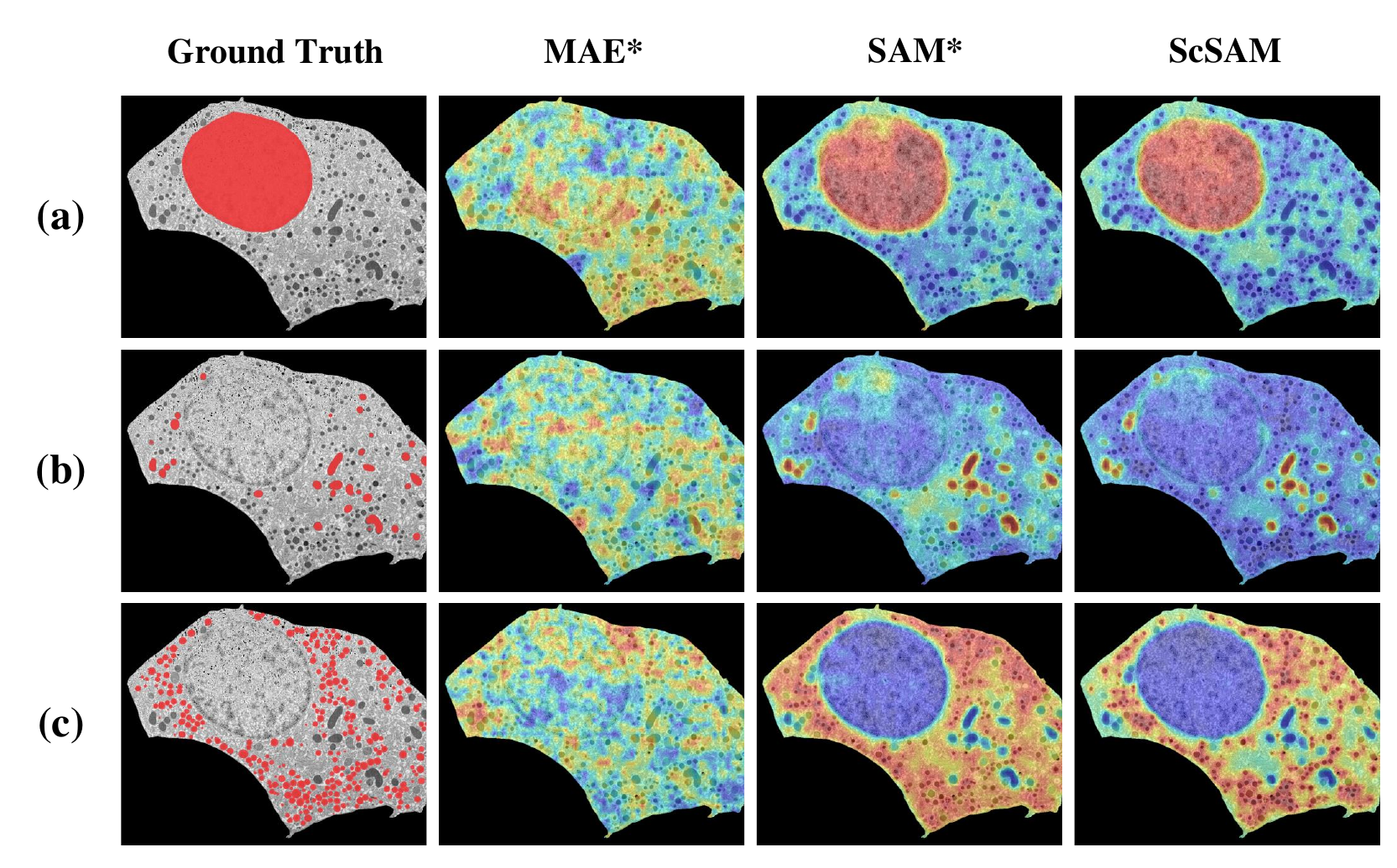} % 插入图片，指定宽度为页面宽度的一半
	\caption{Visualization of positive class similarity maps for three categories: (a) nucleus; (b) mitochondria; (c) granules. We evaluate the cosine similarity between image embeddings and trained class prototypes on a pixel-by-pixel basis and overlay the similarity matrix over the original EMI for comparison.} % 图片说明
	\label{fig:pc} % 图片标签，用于引用
\end{figure}
\subsection{Datasets and Evaluation}
We validate ScSAM on an islet cell dataset called BetaSeg \cite{muller2021jcb}, which consists of two subsets available via OpenOrganelle \cite{heinrich2021nature}.
These two datasets are islet samples isolated under high- and low-glucose conditions and acquired Focused Ion Beam Scanning Electron Microscope (FIB-SEM).
The high- and low-glucose BetaSeg contains three and four islet cells possessing paired reference annotations.
Since manual annotations are only provided for the target cells in EMIs, and the functional relationship between the nucleus and neighboring organelles is crucial for analyzing cellular behavior, we select the middle 350 slices from each islet cell to form our dataset.

Each cell contains binary masks for seven subcells: nucleus, mitochondria, granules, centriole, golgi, membrane, and microtubule.
Among them, we choose the nucleus, mitochondria, and granules, which are crucial for normal cellular function and widely distributed, as segmentation targets.
In addition, to evaluate the performance of ScSAM in a limited sample learning scenario, the last cell of the two datasets serves as the training set, and ScSAM is validated on the remaining cells.
We choose the commonly employed Challenge IoU, mean IoU (m IoU), and Dice score as reference metrics for evaluation.
All comparison and ablation experiments are conducted utilizing the same training strategy and evaluation metrics.

Due to the significant resolution variations and lack of precise labels in EMIs, we choose appropriate preprocessing and augmentation strategies.
Specifically, we select intermediate slices to reduce the biased learning of unlabeled cells and normalize all EMIs.
For data augmentation, the original frames are randomly cropped and resized to enhance edge learning and reduce morphological gaps.

\subsection{Implementation Details}
We apply the same preprocessing pipeline to both the high-glucose and low-glucose subsets of BetaSeg.
For the pre-trained MAE, the field of view (FOV) is set to 256$\times$256, with a patch size 16, converting the FOV into multiple 16$\times$16 patches.
The MAE embedding shape of an EMI slice is 64$\times$64$\times$512, aligning with the size of the SAM embedding.
For islet subcells, the reference prototype dimension for each class is set to 256, while the mask decoder utilizes 128 hidden units. 
The two main experiments employ the Adam optimizer with an initial learning rate of 0.001, while the batch size and epoch are set to 32 and 100 in the training phase.
Notably, we prepare the SAM and MAE embeddings in advance without updating the encoder parameters during training, reducing the computational burden.
The ScSAM is implemented in Pytorch 1.11.0, and all experiments are trained and validated employing an Nvidia GeForce RTX 3090 24GB GPU.

\subsection{Main Results}
We quantitatively compare ScSAM against two groups of methods on both high- and low-glucose BetaSeg datasets (as shown in Tables \ref{tab:cr1} and \ref{tab:cr2}).
The first group includes fully supervised segmentation models, and the second group comprises approaches with frozen pretrained encoders, since no prior work specifically addresses subcellular segmentation under such conditions.
For fair comparisons, those frameworks are conducted with the same processing and training settings as ScSAM.
Notably, frameworks such as MedSAM \cite{ma2024segment} that implement specific target segmentation need to provide precise prompts, which is different from the core mission of ScSAM.

Our supervised baselines include Unet \cite{ronneberger2015u}, AttUnet \cite{oktay2018attention}, nnUnet \cite{isensee2021nnu}, TransUnet \cite{chen2021transunet}, and nnFormer \cite{zhou2021nnformer}, all of which have demonstrated SOTA performance on medical datasets.
For methods that leverage frozen pretrained encoders, we compare the proposed method with the original SAM, including Vanilla SAM \cite{kirillov2023iccv} and Vanilla SAM2 \cite{ravi2024sam2}, which are powerful universal segmentation models, and the SAM-based multi-class semantic segmentation framework, for instance, SurgicalSAM \cite{yue2024aaai}, which is unrestricted to specific application scenarios.
SAM* and MAE* are based on our framework of the class prompt encoder and mask decoder that feed into SAM and MAE embeddings, respectively.
These algorithms train with subcellular slices under the same preprocessing and augmentation strategies for fair comparisons.
In addition, we assign the Vanilla SAM and SAM2 correctly segmented contours to the corresponding organelles and categorize the misrecognized contours to the nearest category.

\begin{table*}[t]
	\centering
	\caption{Comparison results on the high-glucose BetaSeg dataset. The bold figures represent the best performance for each metric.}
	% 设置表格列间距
	\setlength{\tabcolsep}{2mm}{
		\begin{tabular}{l l c c c c c c c}
			\toprule
			% 第一行表头（无加粗）
			\multirow{2}{*}{Training Strategy} & 
			\multirow{2}{*}{Method} & 
			\multirow{2}{*}{Challenge IoU} & 
			\multirow{2}{*}{m IoU} & 
			\multirow{2}{*}{AJI} & 
			\multicolumn{4}{c}{Dice score} \\
			\cmidrule(lr){6-9}
			& & & & & nucleus & mitochondria & granules & overall \\
			\midrule
			
			%========== Specialist Model部分 ==========%
			\multirow{4}{*}{Supervised}
			& Unet      & 0.687 & 0.693 & 0.742 & 0.982 & 0.698 & 0.722 & 0.801 \\
			& AttUnet   & 0.698 & 0.702 & 0.748 & 0.975 & 0.714 & 0.735 & 0.808 \\
			& nnUnet    & 0.701 & 0.704 & 0.756 & 0.981 & 0.747 & 0.732 & 0.821 \\
			& TransUnet & 0.703 & 0.706 & 0.762 & 0.984 & 0.757 & 0.724 & 0.822 \\
			& nnFormer  & 0.716 & 0.720 & 0.772 & 0.981 & 0.774 & 0.745 & 0.833 \\
			
			\midrule
			
			%========== SAM-based Model部分 ==========%
			\multirow{7}{*}{Frozen Encoder}
			& Vanilla SAM         & 0.652 & 0.621 & 0.648 & 0.964 & 0.767 & 0.505 & 0.776 \\
			& Vanilla SAM2        & 0.570 & 0.555 & 0.584 & 0.970 & 0.715 & 0.264 & 0.650 \\
			& SurgicalSAM         & 0.746 & 0.748 & 0.727 & 0.979 & 0.792 & 0.765 & 0.845 \\
			% 在 SurgicalSAM 之后插入一条不完全横线
			& SAM*                & 0.733 & 0.735 & 0.747 & 0.985 & 0.789 & 0.729 & 0.834 \\
			& MAE*                & 0.034 & 0.075 & 0.122 & 0.081 & 0.022 & 0.063 & 0.055 \\
			\cmidrule(lr){2-9}
			& ScSAM (w/o FAFM)    & 0.754 & 0.756 & 0.764 & 0.984 & 0.803 & 0.763 & 0.850 \\
			& ScSAM               & \textbf{0.783} & \textbf{0.785} & \textbf{0.799} 
			& \textbf{0.986} & \textbf{0.830} 
			& \textbf{0.798} & \textbf{0.866} \\
			\bottomrule
		\end{tabular}
	}
	\label{tab:cr1}
\end{table*}

\begin{table*}[t]
	\centering
	\caption{Comparison results on the low-glucose BetaSeg dataset. The bold figures represent the best performance for each metric.}
	% 与第一张表统一: 列间距设置为 2.5mm
	\setlength{\tabcolsep}{2mm}{
		\begin{tabular}{l l c c c c c c c}
			\toprule
			%============= 表头，与第一张表风格保持一致 =============%
			\multirow{2}{*}{Training Strategy} & 
			\multirow{2}{*}{Method} & 
			\multirow{2}{*}{Challenge IoU} & 
			\multirow{2}{*}{m IoU} & 
			\multirow{2}{*}{AJI} & 
			\multicolumn{4}{c}{Dice score} \\
			\cmidrule(lr){6-9}
			& & & & & nucleus & mitochondria & granules & overall \\
			\midrule
			
			%============== Supervised Learning 部分（示例） ==============%
			% 如果实际只有 Frozen Encoder 方法，可删除此区块
			\multirow{4}{*}{Supervised}
			& Unet      & 0.642 & 0.644 & 0.669 & 0.944 & 0.743 & 0.590 & 0.759 \\
			& AttUnet   & 0.646 & 0.649 & 0.683 & 0.953 & 0.728 & 0.614 & 0.765 \\
			& nnUnet    & 0.660 & 0.662 & 0.683 & 0.954 & 0.747 & 0.625 & 0.775 \\
			& TransUnet & 0.663 & 0.668 & 0.685 & 0.962 & 0.772 & 0.597 & 0.777 \\
			& nnFormer  & 0.672 & 0.674 & 0.702 & 0.968 & 0.768 & 0.632 & 0.789 \\
			\midrule
			
			%============== Frozen Encoder 部分 ==============%
			\multirow{7}{*}{Frozen Encoder}
			& Vanilla SAM   & 0.618 & 0.584 & 0.615 & 0.965 & 0.769 & 0.385 & 0.706 \\
			& Vanilla SAM2  & 0.556 & 0.518 & 0.542 & 0.966 & 0.694 & 0.224 & 0.628 \\
			& SurgicalSAM   & 0.744 & 0.746 & 0.739 & 0.976 & 0.807 & 0.741 & 0.842 \\
			% 在 SurgicalSAM 之后插入不完全横线
			& SAM*          & 0.743 & 0.766 & 0.743 & 0.973 & 0.854 & 0.748 & 0.858 \\
			& MAE*          & 0.047 & 0.077 & 0.088 & 0.173 & 0.031 & 0.053 & 0.086 \\
			\cmidrule(lr){2-9}
			& ScSAM (w/o FAFM) & 0.720 & 0.707 & 0.724 & 0.931 & 0.810 & 0.730 & 0.824 \\
			& ScSAM           & \textbf{0.785} & \textbf{0.787} & \textbf{0.798} & \textbf{0.977} 
			& \textbf{0.873} & \textbf{0.767} & \textbf{0.872} \\
			\bottomrule
		\end{tabular}
	}
	\label{tab:cr2}
\end{table*}
\begin{table*}[t]
	\centering
	\caption{Comparison of ScSAM components. Fuse, $L_{cos}$, Dens, and Spar correspond to fusion modules, alignment modules, and dense and sparse embeddings, respectively. Dice\textsubscript{gra} is the Dice score of granules and C IoU stands for Challenge IoU.} 
	\small
	\begin{tabular}{cccc||cccc|cccc}
		\toprule
		\multicolumn{4}{c}{Components} & \multicolumn{4}{c}{High-glucose} & \multicolumn{4}{c}{Low-glucose} \\
		\cmidrule(lr){1-4}\cmidrule(lr){5-8} \cmidrule(lr){9-12}
		Fuse & $L_{cos}$ & Dens & Spar & C IoU & AJI & m Dice & Dice\textsubscript{gra} & C IoU & AJI & m Dice & Dice\textsubscript{gra}\\
		\midrule
		$\checkmark$ & &  $\checkmark$ &  $\checkmark$ & 0.754 & 0.681 & 0.850 & 0.763 & 0.720 & 0.647 & 0.824 & 0.730\\
		$\checkmark$ & $\checkmark$  & & $\checkmark$ & 0.192 & 0.154 & 0.296 &0.230 & 0.191 & 0.132 & 0.234 & 0.214\\
		$\checkmark$ & $\checkmark$  & $\checkmark$ & & 0.763 & 0.697 & 0.856 & 0.775 & 0.771 & 0.673 & 0.864 & 0.752 \\
		\midrule
		$\checkmark$ & $\checkmark$ & $\checkmark$ & $\checkmark$ & \textbf{0.783} & \textbf{0.799} & \textbf{0.866} & \textbf{0.798} & \textbf{0.785} & \textbf{0.798} & \textbf{0.872} & \textbf{0.767}\\
		\bottomrule
	\end{tabular}
	\label{tab:abla}
\end{table*}

Specifically, ScSAM delivers the top score on every aggregate metric (Challenge IoU, m IoU, AJI, overall Dice) in both nutritional settings, as indicated by bolded figures in Tables \ref{tab:cr1} and \ref{tab:cr2}.
In particular, the m IoU improves by 11.3 \% in low-glucose scenarios, demonstrating excellent cross-domain robustness.
As shown in Fig.~\ref{fig:results}, the supervised model performs well for the nucleus but does not accurately depict small structures such as mitochondria and granules.
In contrast, ScSAM improves the Dice scores for mitochondria and granule segmentation to 0.830 and 0.798 in high-glucose, and to 0.873 and 0.767 in low-glucose, indicating that the class prompt encoder and FAFM are effective in alleviating the issues of class imbalance and detail loss.

In terms of SAM-based methods, those frameworks exhibit inferior performance, especially for granule recognition, making it challenging to cope with complex data distributions as they ignore local detail information and fine-grained features.
Vanilla SAM and SAM2 ignore local texture cues, yielding granule Dice scores of just 0.505 and 0.224, respectively.
SurgicalSAM uses only a single feature embedding type, fails to capture fine-grained subcellular variation, and lags behind ScSAM by 3-5\% on all primary metrics.
Moreover, Fig.~\ref{fig:results} exhibits qualitative results for SAM-based approaches on islet-cell segmentation.
ScSAM outlines precise contours across both glucose conditions, revealing superior boundary integrity and strong robustness to domain shifts.

In addition, ScSAM can distinctly differentiate between the three organelles and excels in edge detection of mitochondria and granules, providing an essential prerequisite for disease research and diagnosis.
When applied to subcellular scenarios, this framework updates a small number of parameters and requires limited labeled EMIs, significantly reducing the computational burden while maintaining high efficiency.
ScSAM updates 27.6M parameters during experimental training, achieving optimal performance within 50 epochs.
\begin{figure}[t]
	\centering
	\hspace*{\fill}
	\hspace*{\fill}
	\subfigure[]{
		\includegraphics[width=0.2\textwidth]{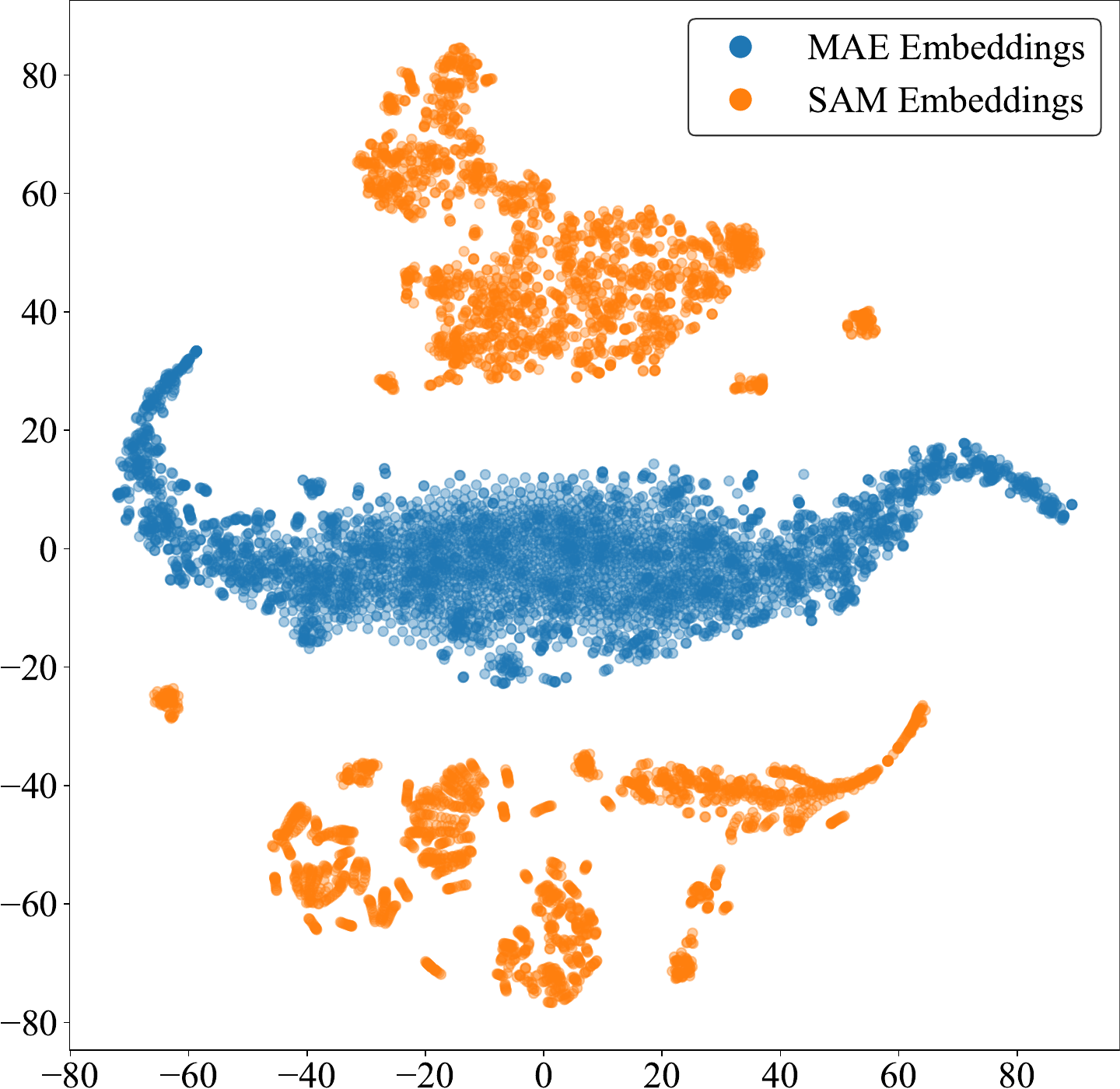}
		\label{fig:tsne_ori}
	}
	\hfill
	\subfigure[]{
		\includegraphics[width=0.2\textwidth]{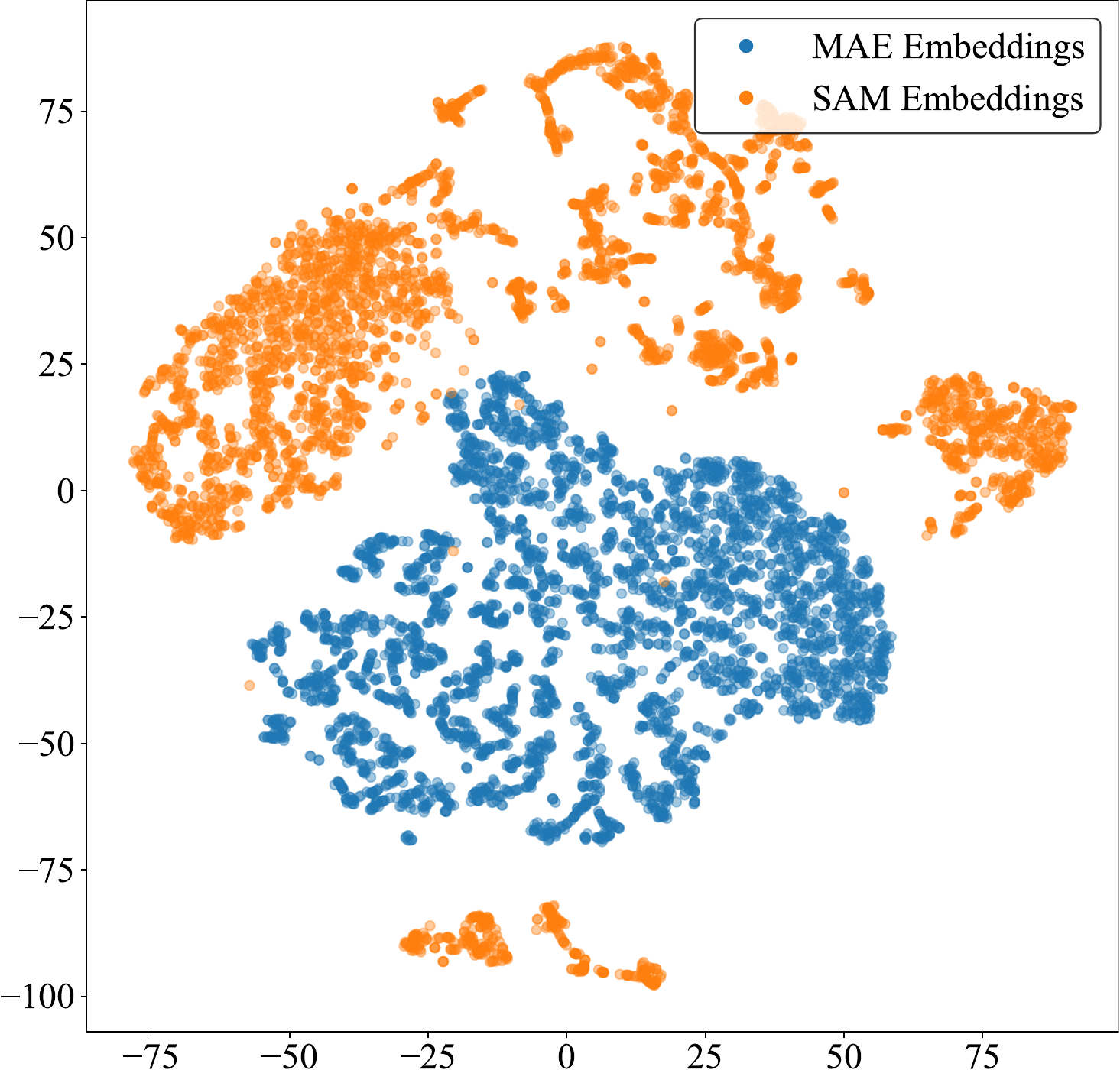}
		\label{fig:tsne_ali}
	}
	\hspace*{\fill}
	\hspace*{\fill}
	\caption{t-SNE discrete visualization embedding distribution for the high-glucose dataset: (a) Original distributions; (b) Aligned distributions.}
	\label{fig:tsne}
\end{figure}
\subsection{Ablation Study}
Tables \ref{tab:cr1} and \ref{tab:cr2} list the results of the ablation study conducted on BetaSeg datasets.
Notably, we remove each component individually to demonstrate its effectiveness as ScSAM's components are tightly coupled.
It can be observed that MAE embeddings trained for image reconstruction are challenging to directly employ for segmentation tasks due to the lack of high-level semantic information.
SAM embeddings conflate visually similar minority classes (mitochondria and granules), leading to lower AJI due to the lack of ultrastructural diversity in their training distributions under the electron microscope.
However, directly concatenating and downscaling two embeddings according to \cite{zhang2023miccai} is even lower than a single input for subcellular recognition due to feature space inconsistency and ineffective fusion strategies.
Moreover, Fig.~\ref{fig:pc} visualizes the pixel-level positive class similarity maps of the individual and dual encoders, exhibiting the precise contours of mitochondria and granules, demonstrating robustness in the long-tailed small target scenario.

Table \ref{tab:abla} presents the results of the ablation experiments, demonstrating the effectiveness of key components and fused representations.
Sparse embeddings have strong category semantics over dense embeddings, but the lack of geometric detail prevents them from distinguishing neighboring contours, leading to poor segmentation predictions.
Overall, each component is essential and maintains a positive effect on subcellular recognition.

In addition, we utilize t-SNE \cite{van2008visualizing} to visualize the scatterplots of the original and aligned embeddings for qualitative analysis of the feature spatial distribution.
As illustrated in Fig.~\ref{fig:tsne}, SAM and MAE embeddings are separated in t-SNE space with no overlap, indicating that the representations differ significantly in feature space within the original distribution.
The aligned embeddings have significant mixing and overlapping in the t-SNE space, demonstrating that the FAFM maps the representations from different sources into a common feature space.

\begin{table}[t]
	\centering
	\caption{Cross-dataset generalization validation. \textit{T}, \textit{V}, \textit{H}, and \textit{L} represent the training and validation sets and the high- and low-glucose BetaSeg datasets, respectively.}
	\scriptsize
	\resizebox{\columnwidth}{!}{
		\hspace{6pt}
		\begin{tabular}{cccccccc}
			\toprule
			\multirow{2}{*}{\textit{T}} & \multirow{2}{*}{\textit{V}} & \multirow{2}{*}{Method} & \multicolumn{3}{c}{Organelles (Dice)} & \multirow{2}{*}{Mean Dice} \\
			\cmidrule{4-6} 
			& & & nuc & mit & gra  & \\
			\midrule
			\multirow{2}{*}{\textit{H}} & \multirow{2}{*}{\textit{L}} & SAM* & 0.929 & 0.727 & 0.679 & 0.778 \\
			& & ScSAM & \textbf{0.967} & \textbf{0.798} & \textbf{0.722} & \textbf{0.829} \\
			\midrule
			\multirow{2}{*}{\textit{L}} & \multirow{2}{*}{\textit{H}} & SAM* & 0.955 & 0.733 & 0.715 & 0.801 \\
			& & ScSAM & \textbf{0.983} & \textbf{0.758} & \textbf{0.783} & \textbf{0.833}  \\
			\bottomrule
		\end{tabular}
	\hspace{6pt}
	}
	\label{tab:example}
\end{table}

\subsection{Efficiency Comparisons}
To comprehensively evaluate the training and inference efficiency of ScSAM, we present the time costs of different models under the same device environment and settings in Table \ref{tab:ctr}.
Although ScSAM is devised with a dual-encoder architecture, its inference time is only 0.457 s per image, which is 0.14 s slower than that of Vanilla SAM, yet it still meets the throughput requirements for offline cell behavior analysis.
MAE* achieves the fastest inference speed of 0.144 s per image, but its backbone lacks advanced semantic priors, resulting in suboptimal contour recognition.

\begin{figure}[t]
	\centering
	\hspace*{\fill}
	\subfigure[]{
		\includegraphics[width=0.21\textwidth]{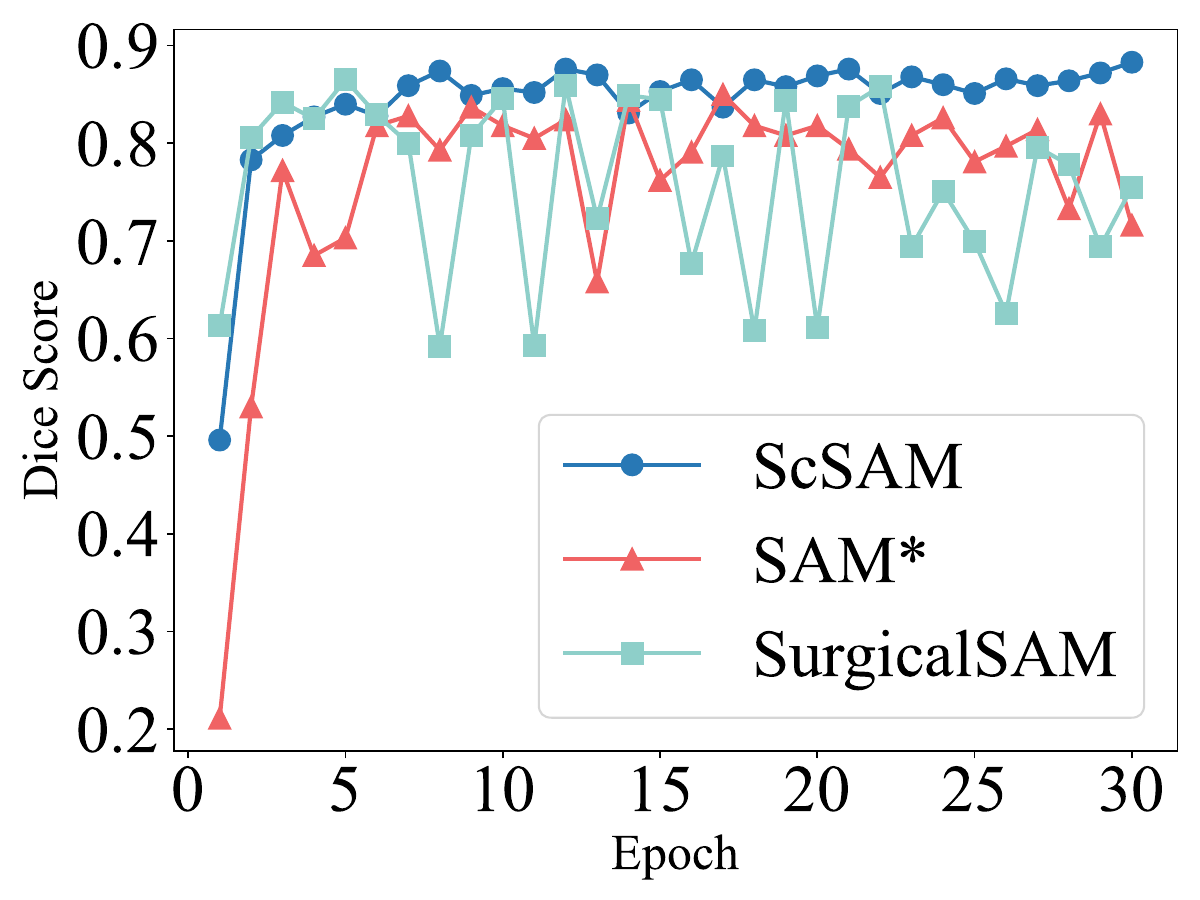}
	}
	\hfill
	\subfigure[]{
		\includegraphics[width=0.21\textwidth]{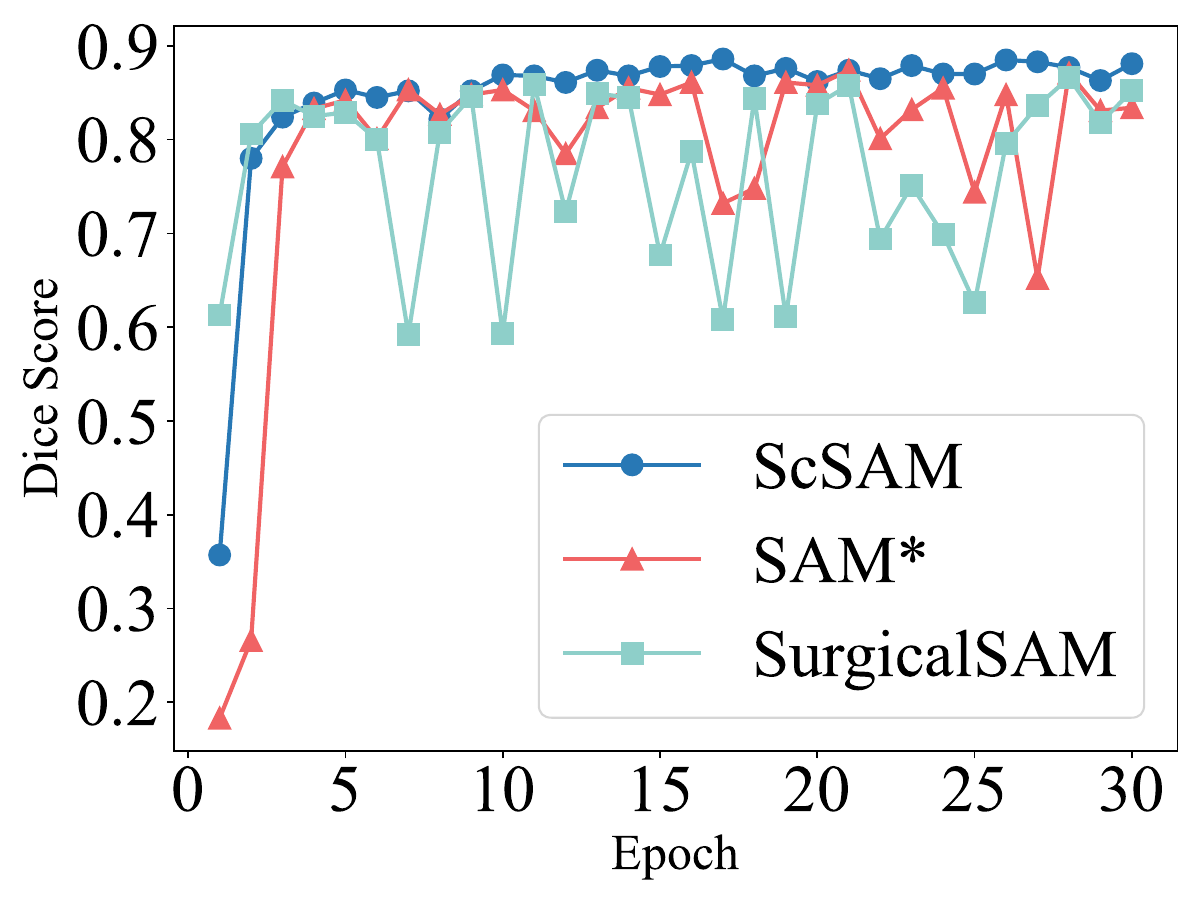}
	}
	\hspace*{\fill}
	\caption{Dice score over 30 epochs for comparison methods trained on high- and low-glucose BetaSeg datasets. We compare robustness and convergence speed by visualizing the evolution of the mean Dice score.}
	\label{fig:dice}
\end{figure}

Although dual backbones typically incur higher computational cost, ScSAM exhibits excellent training efficiency, reaching peak Dice scores in just 3.2 hours, compared to 6.5 hours for SurgicalSAM, 4.6 hours for MAE*, and 4.7 hours for SAM*.
Two factors underlie the rapid convergence.
First, the SAM and MAE backbones are frozen, while only the lightweight FAFM and class prompt encoder require updated parameters, which shrinks the optimization exploration space.
Second, the contrastive loss expands the interclass angular boundaries and tightens the intraclass clustering, increasing the initial gradient and reducing the variance, thus accelerating the model convergence.
Fig.~\ref{fig:dice} presents the learning curves for the initial thirty epochs, where ScSAM has a higher Dice score and exhibits greater stability than SurgicalSAM and SAM*.

$  $

\subsection{Cross-Dataset Generalization}

We further evaluate ScSAM's robustness and transferability for subcellular recognition by training on one dataset and testing on another.
As summarized in Table \ref{tab:example}, we compared Dice scores for single-encoder inputs (SAM* baseline) and our cross-task fusion input under two asymmetric settings: high-to-low glucose and low-to-high glucose.
In both directions, ScSAM consistently surpasses the SAM* baseline, achieving an absolute Dice improvement of 5.3 \% for granule segmentation and maintaining stable performance on nucleus and mitochondria classes.

Notably, ScSAM depicts accurate contours of granules even when confronted with islet cells that exhibit strong compositional contrast, demonstrating its ability to disentangle morphology from imaging contrast.
These inspiring results indicate that cross-task fusion representations maintain robust performance in cell images with remarkable contrast and culture environment differences, balancing domain shift across contrasts and capturing domain-invariant features.
By leveraging the complementary structural cues of MAE and the high-level semantics of SAM, SCSAM balances intensity-driven contrast variations and preserves discriminative subcellular patterns across datasets with different staining protocols and nutrient levels.

\begin{table}[t]
	\centering
	\caption{Comparison results of training and inference time}
	\small % 或 \scriptsize
	\setlength{\tabcolsep}{4pt} % 调整列间距
	\begin{tabular}{lcc}
		\toprule
		Method & Training time (h) & Inference time (s)\\
		\midrule
		Vanilla SAM  & -   & 0.321\\
		Vanilla SAM2 & -   & 0.273\\
		SurgicalSAM  & 6.5 & 0.374\\
		MAE*         & 4.6 & 0.144\\
		SAM*         & 4.7 & 0.375\\
		ScSAM        & 3.2 & 0.457\\
		\bottomrule
	\end{tabular}
	\label{tab:ctr}
\end{table}
\section{Conclusion}
In this paper, we introduce ScSAM, a cross-task alignment and fusion framework that alleviates the morphological and distributional biases hampering subcellular semantic segmentation.
ScSAM first employs a lightweight Feature-Alignment and Fusion Module (FAFM) to map heterogeneous SAM and MAE embeddings into a shared latent space and fuse them adaptively, thereby maximizing complementary information.
To substitute for the explicit prompts, we construct a cosine similarity matrix in the class prompt encoder and employ contrastive learning loss to generate relevant prompts to activate class-aware regions while suppressing extraneous information expression.
Experiments on diverse electron-microscopy datasets demonstrate that ScSAM tackles the morphological and distributional shifts existing in subcellular recognition.
In future work, we will extend our cross-task fusion strategy to volume electron microscopy and other biomedical domains with resolution and class-imbalance shifts.
%%%%%%%%%%%%%%%%%%%%%%%%%%%%%%%%%%%%%%%%%%%%%%%%%%%%%%%%%%%%%%%%%%%%%%%%

%%% Use this environment to include acknowledgements (optional).
%%% This will be omitted in doubleblind mode.

%%%%%%%%%%%%%%%%%%%%%%%%%%%%%%%%%%%%%%%%%%%%%%%%%%%%%%%%%%%%%%%%%%%%%%%%

%%% Use this command to include your bibliography file.
\bibliographystyle{ecai}  
\bibliography{mybibfile}

\begin{thebibliography}{42}
\providecommand{\natexlab}[1]{#1}
\providecommand{\url}[1]{\texttt{#1}}
\expandafter\ifx\csname urlstyle\endcsname\relax
  \providecommand{\doi}[1]{doi: #1}\else
  \providecommand{\doi}{doi: \begingroup \urlstyle{rm}\Url}\fi

\bibitem[Abdur~Rahman et~al.(2024)Abdur~Rahman, Papathanail, Brigato, and
  Mougiakakou]{abdur2024sam}
L.~Abdur~Rahman, I.~Papathanail, L.~Brigato, and S.~Mougiakakou.
\newblock A {SAM} based tool for semi-automatic food annotation.
\newblock In \emph{Proceedings of the European Conference on Artificial
  Intelligence}, pages 4475--4478. IOS Press, 2024.

\bibitem[Bissoto et~al.(2020)]{bissoto2020debiasing}
A.~Bissoto et~al.
\newblock Debiasing skin lesion datasets and models? not so fast.
\newblock In \emph{Proceedings of the IEEE/CVF Conference on Computer Vision
  and Pattern Recognition Workshops}, pages 740--741, 2020.

\bibitem[Boykov et~al.(2006)]{boykov2006graph}
Y.~Boykov et~al.
\newblock Graph cuts and efficient {N-D} image segmentation.
\newblock \emph{International Journal of Computer Vision}, 70\penalty0
  (2):\penalty0 109--131, 2006.

\bibitem[Chan and Vese(2001)]{chan2001active}
T.~F. Chan and L.~A. Vese.
\newblock Active contours without edges.
\newblock \emph{IEEE Transactions on Image Processing}, 10\penalty0
  (2):\penalty0 266--277, 2001.

\bibitem[Chen et~al.(2021)Chen, Lu, Yu, Luo, Adeli, Wang, Lu, Yuille, and
  Zhou]{chen2021transunet}
J.~Chen, Y.~Lu, Q.~Yu, X.~Luo, E.~Adeli, Y.~Wang, L.~Lu, A.~L. Yuille, and
  Y.~Zhou.
\newblock {TransUnet}: Transformers make strong encoders for medical image
  segmentation.
\newblock \emph{arXiv preprint arXiv:2102.04306}, 2021.

\bibitem[Chen et~al.(2023)Chen, Zhu, Deng, Cao, Wang, Zhang, Li, Sun, Zang, and
  Mao]{chen2023iccv}
T.~Chen, L.~Zhu, C.~Deng, R.~Cao, Y.~Wang, S.~Zhang, Z.~Li, L.~Sun, Y.~Zang,
  and P.~Mao.
\newblock {SAM-Adapter}: Adapting segment anything in underperformed scenes.
\newblock In \emph{Proceedings of the IEEE/CVF International Conference on
  Computer Vision Workshops}, pages 3367--3375, 2023.

\bibitem[Cheng et~al.(2024)Cheng, Wei, Zhu, Wang, Qu, Shao, and
  Zhou]{cheng2024unleashing}
Z.~Cheng, Q.~Wei, H.~Zhu, Y.~Wang, L.~Qu, W.~Shao, and Y.~Zhou.
\newblock Unleashing the potential of {SAM} for medical adaptation via
  hierarchical decoding.
\newblock In \emph{Proceedings of the IEEE/CVF Conference on Computer Vision
  and Pattern Recognition}, pages 3511--3522, 2024.

\bibitem[Fan et~al.(2023)Fan, Liu, Chang, Huang, Chen, and
  Cai]{fan2023taxonomy}
J.~Fan, D.~Liu, H.~Chang, H.~Huang, M.~Chen, and W.~Cai.
\newblock Taxonomy adaptive cross-domain adaptation in medical imaging via
  optimization trajectory distillation.
\newblock In \emph{Proceedings of the IEEE/CVF International Conference on
  Computer Vision}, pages 21174--21184, 2023.

\bibitem[Fan et~al.(2024{\natexlab{a}})Fan, Liu, Chang, Huang, Chen, and
  Cai]{fan2024seeing}
J.~Fan, D.~Liu, H.~Chang, H.~Huang, M.~Chen, and W.~Cai.
\newblock Seeing unseen: Discover novel biomedical concepts via
  geometry-constrained probabilistic modeling.
\newblock In \emph{Proceedings of the IEEE/CVF Conference on Computer Vision
  and Pattern Recognition}, pages 11524--11534, 2024{\natexlab{a}}.

\bibitem[Fan et~al.(2024{\natexlab{b}})Fan, Liu, Li, Chang, Huang, Braet, Chen,
  and Cai]{fan2024revisiting}
J.~Fan, D.~Liu, C.~Li, H.~Chang, H.~Huang, F.~Braet, M.~Chen, and W.~Cai.
\newblock Revisiting adaptive cellular recognition under domain shifts: A
  contextual correspondence view.
\newblock In \emph{European Conference on Computer Vision}, pages 275--292.
  Springer, 2024{\natexlab{b}}.

\bibitem[Han et~al.(2022)]{han2022cvpr}
H.~Han et~al.
\newblock Self-supervised voxel-level representation rediscovers subcellular
  structures in volume electron microscopy.
\newblock In \emph{Proceedings of the IEEE/CVF Conference on Computer Vision
  and Pattern Recognition}, pages 1874--1883, 2022.

\bibitem[He et~al.(2023)He, Unberath, Ke, and Shen]{he2023miccai}
Z.~He, M.~Unberath, J.~Ke, and Y.~Shen.
\newblock {TransNuSeg}: A lightweight multi-task transformer for nuclei
  segmentation.
\newblock In \emph{Medical Image Computing and Computer Assisted Intervention
  -- MICCAI 2023}, pages 206--215. Springer, 2023.

\bibitem[Heinrich et~al.(2021)Heinrich, Bennett, Ackerman, Park, Bogovic,
  Eckstein, Petruncio, Clements, Pang, Xu, et~al.]{heinrich2021nature}
L.~Heinrich, D.~Bennett, D.~Ackerman, W.~Park, J.~Bogovic, N.~Eckstein,
  A.~Petruncio, J.~Clements, S.~Pang, C.~S. Xu, et~al.
\newblock Whole-cell organelle segmentation in volume electron microscopy.
\newblock \emph{Nature}, 599\penalty0 (7883):\penalty0 141--146, 2021.

\bibitem[Isensee et~al.(2021)Isensee, Jaeger, Kohl, Petersen, and
  Maier-Hein]{isensee2021nnu}
F.~Isensee, P.~F. Jaeger, S.~A. Kohl, J.~Petersen, and K.~H. Maier-Hein.
\newblock {nnU-Net}: a self-configuring method for deep learning-based
  biomedical image segmentation.
\newblock \emph{Nature Methods}, 18\penalty0 (2):\penalty0 203--211, 2021.

\bibitem[Kirillov et~al.(2023)Kirillov, Mintun, Ravi, Mao, Rolland, Gustafson,
  Xiao, Whitehead, Berg, Lo, et~al.]{kirillov2023iccv}
A.~Kirillov, E.~Mintun, N.~Ravi, H.~Mao, C.~Rolland, L.~Gustafson, T.~Xiao,
  S.~Whitehead, A.~C. Berg, W.-Y. Lo, et~al.
\newblock Segment anything.
\newblock In \emph{Proceedings of the IEEE/CVF International Conference on
  Computer Vision}, pages 4015--4026, 2023.

\bibitem[Li et~al.(2023)]{li2023cvpr}
X.~Li et~al.
\newblock {LoGoNet}: Towards accurate {3D} object detection with
  local-to-global cross-modal fusion.
\newblock In \emph{Proceedings of the IEEE/CVF Conference on Computer Vision
  and Pattern Recognition}, pages 17524--17534, 2023.

\bibitem[Lin et~al.(2024)Lin, Liu, Lu, and Jia]{lin2024cvpr}
J.~Lin, L.~Liu, D.~Lu, and K.~Jia.
\newblock {SAM-6D}: Segment anything model meets zero-shot {6D} object pose
  estimation.
\newblock In \emph{Proceedings of the IEEE/CVF Conference on Computer Vision
  and Pattern Recognition}, pages 27906--27916, 2024.

\bibitem[Liu et~al.(2020)]{liu2020pdam}
D.~Liu et~al.
\newblock {PDAM}: A panoptic-level feature alignment framework for unsupervised
  domain adaptive instance segmentation in microscopy images.
\newblock \emph{IEEE Transactions on Medical Imaging}, 40\penalty0
  (1):\penalty0 154--165, 2020.

\bibitem[Luo et~al.(2024)Luo, Sun, Pan, Zhang, and Wu]{luo2024aaai}
N.~Luo, R.~Sun, Y.~Pan, T.~Zhang, and F.~Wu.
\newblock Electron microscopy images as set of fragments for mitochondrial
  segmentation.
\newblock In \emph{Proceedings of the AAAI Conference on Artificial
  Intelligence}, volume~38, pages 3981--3989, 2024.

\bibitem[Ma et~al.(2024)Ma, He, Li, Han, You, and Wang]{ma2024segment}
J.~Ma, Y.~He, F.~Li, L.~Han, C.~You, and B.~Wang.
\newblock Segment anything in medical images.
\newblock \emph{Nature Communications}, 15\penalty0 (1):\penalty0 654, 2024.

\bibitem[M{\"u}ller et~al.(2021)M{\"u}ller, Schmidt, Xu, Pang, {D{'}Costa},
  Kretschmar, M{\"u}nster, Kurth, Jug, Weigert, et~al.]{muller2021jcb}
A.~M{\"u}ller, D.~Schmidt, C.~S. Xu, S.~Pang, J.~V. {D{'}Costa}, S.~Kretschmar,
  C.~M{\"u}nster, T.~Kurth, F.~Jug, M.~Weigert, et~al.
\newblock {3D FIB-SEM} reconstruction of microtubule--organelle interaction in
  whole primary mouse $\beta$ cells.
\newblock \emph{Journal of Cell Biology}, 220\penalty0 (2), 2021.

\bibitem[Oktay et~al.(2018)]{oktay2018attention}
O.~Oktay et~al.
\newblock Attention {U-net}: Learning where to look for the pancreas.
\newblock \emph{arXiv preprint arXiv:1804.03999}, 2018.

\bibitem[Otsu(1975)]{otsu1975automatica}
N.~Otsu.
\newblock A threshold selection method from gray-level histograms.
\newblock \emph{Automatica}, 11\penalty0 (285-296):\penalty0 23--27, 1975.

\bibitem[Peng et~al.(2024)Peng, Xu, Zeng, Yang, and Shen]{peng2024sam}
Z.~Peng, Z.~Xu, Z.~Zeng, X.~Yang, and W.~Shen.
\newblock {SAM-PARSER}: Fine-tuning {SAM} efficiently by parameter space
  reconstruction.
\newblock In \emph{Proceedings of the AAAI Conference on Artificial
  Intelligence}, volume~38, pages 4515--4523, 2024.

\bibitem[Ravi et~al.(2024)]{ravi2024sam2}
N.~Ravi et~al.
\newblock {SAM} 2: Segment anything in images and videos.
\newblock \emph{arXiv preprint arXiv:2408.00714}, 2024.

\bibitem[Rizk et~al.(2014)Rizk, Paul, Incardona, Bugarski, Mansouri, Niemann,
  Ziegler, Berger, and Sbalzarini]{rizk2014nature}
A.~Rizk, G.~Paul, P.~Incardona, M.~Bugarski, M.~Mansouri, A.~Niemann,
  U.~Ziegler, P.~Berger, and I.~F. Sbalzarini.
\newblock Segmentation and quantification of subcellular structures in
  fluorescence microscopy images using squassh.
\newblock \emph{Nature Protocols}, 9\penalty0 (3):\penalty0 586--596, 2014.

\bibitem[Ronneberger et~al.(2015)Ronneberger, Fischer, and
  Brox]{ronneberger2015u}
O.~Ronneberger, P.~Fischer, and T.~Brox.
\newblock U-net: Convolutional networks for biomedical image segmentation.
\newblock In \emph{Medical Image Computing and Computer-Assisted Intervention
  -- MICCAI 2015}, pages 234--241. Springer, 2015.

\bibitem[Sekh et~al.(2021)Sekh, Opstad, Godtliebsen, Birgisdottir, Ahluwalia,
  Agarwal, and Prasad]{sekh2021nature}
A.~A. Sekh, I.~S. Opstad, G.~Godtliebsen, {\AA}.~B. Birgisdottir, B.~S.
  Ahluwalia, K.~Agarwal, and D.~K. Prasad.
\newblock Physics-based machine learning for subcellular segmentation in living
  cells.
\newblock \emph{Nature Machine Intelligence}, 3\penalty0 (12):\penalty0
  1071--1080, 2021.

\bibitem[Song et~al.(2024)Song, Fan, Huang, Chen, and Cai]{song2024cell}
Y.~Song, J.~Fan, H.~Huang, M.~Chen, and W.~Cai.
\newblock Cell as {Point}: One-stage framework for efficient cell tracking.
\newblock \emph{arXiv preprint arXiv:2411.14833}, 2024.

\bibitem[Van~der Maaten and Hinton(2008)]{van2008visualizing}
L.~Van~der Maaten and G.~Hinton.
\newblock Visualizing data using {t-SNE}.
\newblock \emph{Journal of Machine Learning Research}, 9\penalty0 (11), 2008.

\bibitem[Wang et~al.(2019)Wang, Chao, Garg, Hariharan, Campbell, and
  Weinberger]{wang2019pseudo}
Y.~Wang, W.-L. Chao, D.~Garg, B.~Hariharan, M.~Campbell, and K.~Q. Weinberger.
\newblock Pseudo-lidar from visual depth estimation: Bridging the gap in {3D}
  object detection for autonomous driving.
\newblock In \emph{Proceedings of the IEEE/CVF Conference on Computer Vision
  and Pattern Recognition}, pages 8445--8453, 2019.

\bibitem[Woo et~al.(2018)Woo, Park, Lee, and Kweon]{woo2018cbam}
S.~Woo, J.~Park, J.-Y. Lee, and I.~S. Kweon.
\newblock {CBAM}: Convolutional block attention module.
\newblock In \emph{Proceedings of the European Conference on Computer Vision},
  pages 3--19, 2018.

\bibitem[Xie et~al.(2024)Xie, Guo, Cong, Pagnucco, and Song]{xie2024domain}
K.~Xie, R.~Guo, C.~Cong, M.~Pagnucco, and Y.~Song.
\newblock Domain generalised cell nuclei segmentation in histopathology images
  using domain-aware curriculum learning and colour-perceived meta learning.
\newblock In \emph{Proceedings of the European Conference on Artificial
  Intelligence}, pages 354--361. IOS Press, 2024.

\bibitem[Xie et~al.(2023)Xie, Pang, Bader, and Wang]{xie2023cvpr}
R.~Xie, K.~Pang, G.~D. Bader, and B.~Wang.
\newblock {MAESTER}: Masked autoencoder guided segmentation at pixel resolution
  for accurate, self-supervised subcellular structure recognition.
\newblock In \emph{Proceedings of the IEEE/CVF Conference on Computer Vision
  and Pattern Recognition}, pages 3292--3301, 2023.

\bibitem[Yamagiwa et~al.(2024)]{yamagiwa2024wacv}
H.~Yamagiwa et~al.
\newblock Zero-shot edge detection with {SCESAME}: Spectral clustering-based
  ensemble for segment anything model estimation.
\newblock In \emph{Proceedings of the IEEE/CVF Winter Conference on
  Applications of Computer Vision}, pages 541--551, 2024.

\bibitem[Yuan et~al.(2024)Yuan, Luo, Hui, Pu, Xiang, Ranjan, and
  Demandolx]{yuan2024cvpr}
S.~Yuan, L.~Luo, Z.~Hui, C.~Pu, X.~Xiang, R.~Ranjan, and D.~Demandolx.
\newblock {UnSAMFlow}: Unsupervised optical flow fuided by segment anything
  model.
\newblock In \emph{Proceedings of the IEEE/CVF Conference on Computer Vision
  and Pattern Recognition}, pages 19027--19037, 2024.

\bibitem[Yue et~al.(2024)Yue, Zhang, Hu, Xia, Luo, and Wang]{yue2024aaai}
W.~Yue, J.~Zhang, K.~Hu, Y.~Xia, J.~Luo, and Z.~Wang.
\newblock {SurgicalSAM}: Efficient class promptable surgical instrument
  segmentation.
\newblock In \emph{Proceedings of the AAAI Conference on Artificial
  Intelligence}, volume~38, pages 6890--6898, 2024.

\bibitem[Zhang et~al.(2023)]{zhang2023miccai}
J.~Zhang et~al.
\newblock {SAM-Path}: A segment anything model for semantic segmentation in
  digital pathology.
\newblock In \emph{Medical Image Computing and Computer Assisted Intervention
  -- MICCAI 2023 Workshops}, pages 161--170. Springer, 2023.

\bibitem[Zhang et~al.(2024{\natexlab{a}})Zhang, Yan, Liu, and
  Lu]{zhang2024cvpr}
P.~Zhang, T.~Yan, Y.~Liu, and H.~Lu.
\newblock Fantastic animals and where to find them: Segment any marine animal
  with dual {SAM}.
\newblock In \emph{Proceedings of the IEEE/CVF Conference on Computer Vision
  and Pattern Recognition}, pages 2578--2587, 2024{\natexlab{a}}.

\bibitem[Zhang et~al.(2024{\natexlab{b}})Zhang, Liu, Li, Chen, Liu, Hu, Xiong,
  Yuan, and Wang]{zhang2024distilling}
Q.~Zhang, X.~Liu, W.~Li, H.~Chen, J.~Liu, J.~Hu, Z.~Xiong, C.~Yuan, and
  Y.~Wang.
\newblock Distilling semantic priors from {SAM} to efficient image restoration
  models.
\newblock In \emph{Proceedings of the IEEE/CVF Conference on Computer Vision
  and Pattern Recognition}, pages 25409--25419, 2024{\natexlab{b}}.

\bibitem[Zhou et~al.(2021)Zhou, Guo, Zhang, Yu, Wang, and Yu]{zhou2021nnformer}
H.-Y. Zhou, J.~Guo, Y.~Zhang, L.~Yu, L.~Wang, and Y.~Yu.
\newblock {nnFormer}: Interleaved transformer for volumetric segmentation.
\newblock \emph{arXiv preprint arXiv:2109.03201}, 2021.

\bibitem[Zhou et~al.(2024)Zhou, He, Tan, and Yan]{zhou2024aaai}
S.~Zhou, R.~He, W.~Tan, and B.~Yan.
\newblock {SAMFlow}: Eliminating any fragmentation in optical flow with segment
  anything model.
\newblock In \emph{Proceedings of the AAAI Conference on Artificial
  Intelligence}, volume~38, pages 7695--7703, 2024.

\end{thebibliography}
\clearpage\vfill\section*{Supplementary Material}
\setcounter{section}{0}  
\renewcommand\thefigure{\arabic{figure}}
\setcounter{figure}{0}
\renewcommand\thetable{\arabic{table}}
\setcounter{table}{0}
%%%%%%%%%%%%%%%%%%%%%%%%%%%%%%%%%%%%%%%%%%%%%%%%%%%%%%%%%%%%%%%%%%%%%%%%
\section{Implementation Details}
\subsection{Pre-trained MAE}
MAE is pre-trained on the BetaSeg datasets, utilizing a cropped image size of 256, a patch size of 16, a mask ratio of 0.5, and 1600 epochs.
To generate embeddings, we resize the original EMIs to 1024 pixels on the longest side, then pad them to 1024$\times$1024 with 0 pixels.
The resized image is then fully divided into four pieces and fed into the pre-trained MAE.
The output representations are stitched together according to their original positions to obtain a 64$\times$64$\times$512 embedding, which matches the SAM embedding.
\subsection{Vanilla SAM and SAM2}
Vanilla SAM and SAM2 served as baseline comparison algorithms.
The contours predicted by Vanilla SAM and SAM2 are recognized as the correct category, i.e., the regions where the predicted and real masks overlap are considered to correspond to the true organelles.
In addition, misidentified regions are categorized based on their nearest contour's corresponding organelle.
Notably, both foundation models hardly recognize organelles smaller than the granule size, allowing for a fair comparison under such conditions.
\section{Ablation Study}

\subsection{Cosine Similarity Loss Weights}
We conduct ablation experiments on various cosine similarity loss weights on low- and high-glucose datasets.
Specifically, we compare the segmentation performance of ScSAM with loss weights of 0.1, 0.2, and 0.3, respectively, as shown in Table \ref{tab:asl}.
It can be observed that ScSAM precisely aligns the pre-trained embeddings when the weight of $\mathcal{L}_{cos}$ is 0.2, showing excellent subcellular recognition, especially in tiny organelles such as mitochondria and granules.
We do not report results for a weight of 1 because it severely hinders early-stage model convergence.

\subsection{Residual Connection Structure}
Table \ref{tab:asl1} presents the results of validation experiments conducted on the BetaSeg datasets, demonstrating the effectiveness of the Residual Connection (RC) structure.
ScSAM achieves state-of-the-art performance in each metric, indicating that the residual module significantly enhances the learning of class-specific information within the class prompt encoder, thereby enriching the feature representation in the dense embeddings.

\subsection{Fusion Strategy}
To validate the effectiveness of FAFM, we compare different fusion strategies in Table \ref{tab:asl2}.
Specifically, the concatenation method directly concatenates the MAE and SAM feature vectors in the channel dimension and conducts dimensional alignment by 1 $\times$ 1 convolution.
The cross-attention strategy firstly uses the SAM embedding as Query and the MAE embedding as Key/Value (and then reverses the process to use MAE as Query once more), and utilizes the Query-key correlation to calculate fine-grained weights and adaptively select and fuse complementary information.
As shown in Table \ref{tab:asl2}, by explicitly aligning and adaptively fusing MAE and SAM features, FAFM not only surpasses concatenation strategy but also outperforms symmetric cross-attention.

\begin{table}[t]
	\centering
	\small % 或 \scriptsize
	\setlength{\tabcolsep}{4pt} % 调整列间距
	\begin{tabular}{lcccccc}
		\toprule
		\multirow{2}{*}{$\mathcal{L}_{cos}$ weight} & \multirow{2}{*}{C IoU} & \multirow{2}{*}{AJI} &  \multicolumn{4}{c}{Dice score}\\
		\cmidrule(lr){4-7} 
		&&& nuc & mit & gra & mean\\
		\midrule
		0.1 & 0.773  & 0.781 & 0.981 & 0.836 & 0.775 & 0.864\\
		0.2 (Ours) & \textbf{0.784} & \textbf{0.799} & \textbf{0.982} & \textbf{0.852} & \textbf{0.783} & \textbf{0.869} \\
		0.3 & 0.751 & 0.754 & 0.974 & 0.816 & 0.756 & 0.849\\
		\bottomrule
	\end{tabular}
	\caption{Ablation study of $\mathcal{L}_{cos}$ weight on the BetaSeg dataset.}
	\label{tab:asl}
\end{table}

\begin{table}[h!]
	\centering
	\small % 或 \scriptsize
	\setlength{\tabcolsep}{3.5pt} % 调整列间距
	\begin{tabular}{lcccccc}
		\toprule
		\multirow{2}{*}{Method} & \multirow{2}{*}{C IoU} & \multirow{2}{*}{AJI} &  \multicolumn{4}{c}{Dice score}\\
		\cmidrule(lr){4-7} 
		&&& nuc & mit & gra & mean\\
		\midrule
		ScSAM (w/o RC)  & 0.757 & 0.774 & 0.976 & 0.832 & 0.752 & 0.853\\
		ScSAM & \textbf{0.784} & \textbf{0.799} & \textbf{0.982} & \textbf{0.852} & \textbf{0.783} & \textbf{0.869}\\
		\bottomrule
	\end{tabular}
	\caption{Ablation study of RC on the low-glucose BetaSeg dataset.}
	\label{tab:asl1}
\end{table}

\begin{table}[h!]
	\centering
	\small
	%——开启红色模式——
	\setlength{\tabcolsep}{3.5pt}
	\begin{tabular}{lcccccc}
		\toprule
		\multirow{2}{*}{Method} & \multirow{2}{*}{C IoU} & \multirow{2}{*}{AJI} & \multicolumn{4}{c}{Dice score}\\
		\cmidrule(lr){4-7}
		&&& nuc & mit & gra & mean\\
		\midrule
		Concatenation  & 0.662 & 0.634 & 0.918 & 0.687 & 0.654 & 0.753\\
		Cross-attention & 0.706 & 0.744 & 0.979 & 0.749 & 0.714 & 0.814\\
		FAFM (Ours) & \textbf{0.784} & \textbf{0.799} & \textbf{0.982} & \textbf{0.852} & \textbf{0.783} & \textbf{0.869}\\
		\bottomrule
	\end{tabular}
	%——恢复默认颜色——
	\caption{Ablation study of FAFM on the low-glucose BetaSeg dataset.}
	\label{tab:asl2}
\end{table}

\section{Visualization}
\subsection{Result Visualization}
\begin{figure*}[h!]
	\centering
	\includegraphics[width=0.98\textwidth]{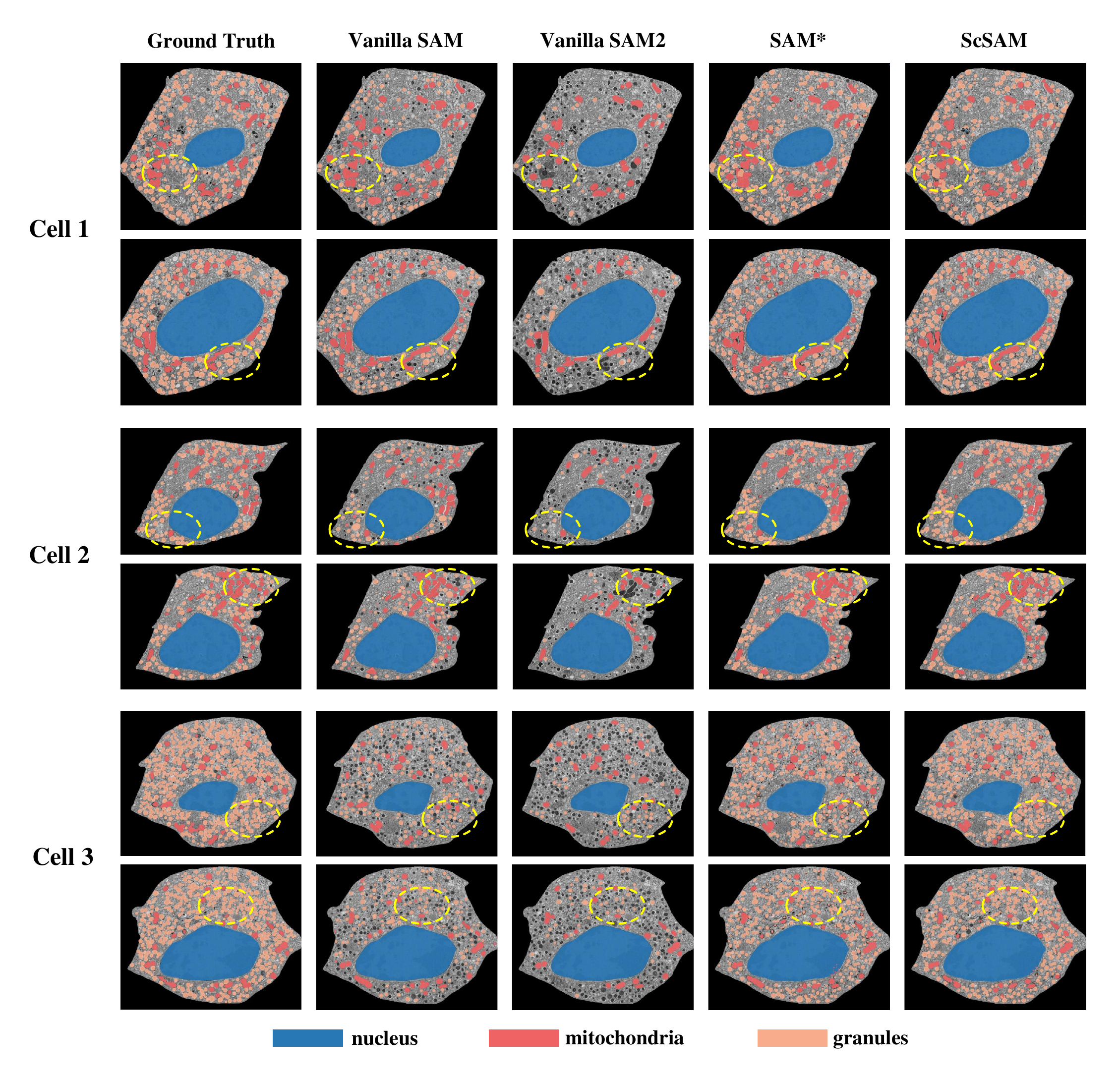}
	\caption{Visual comparison of prediction masks on high-glucose samples. These methods validated one islet cell in each dataset by overlaying the original images and segmentation masks. Notably, the yellow dashed ellipse is used to emphasize regions with significant recognition variance.}	
	\label{fig:result} 
\end{figure*}

\begin{figure*}[h!]
	\centering
	\includegraphics[width=\textwidth]{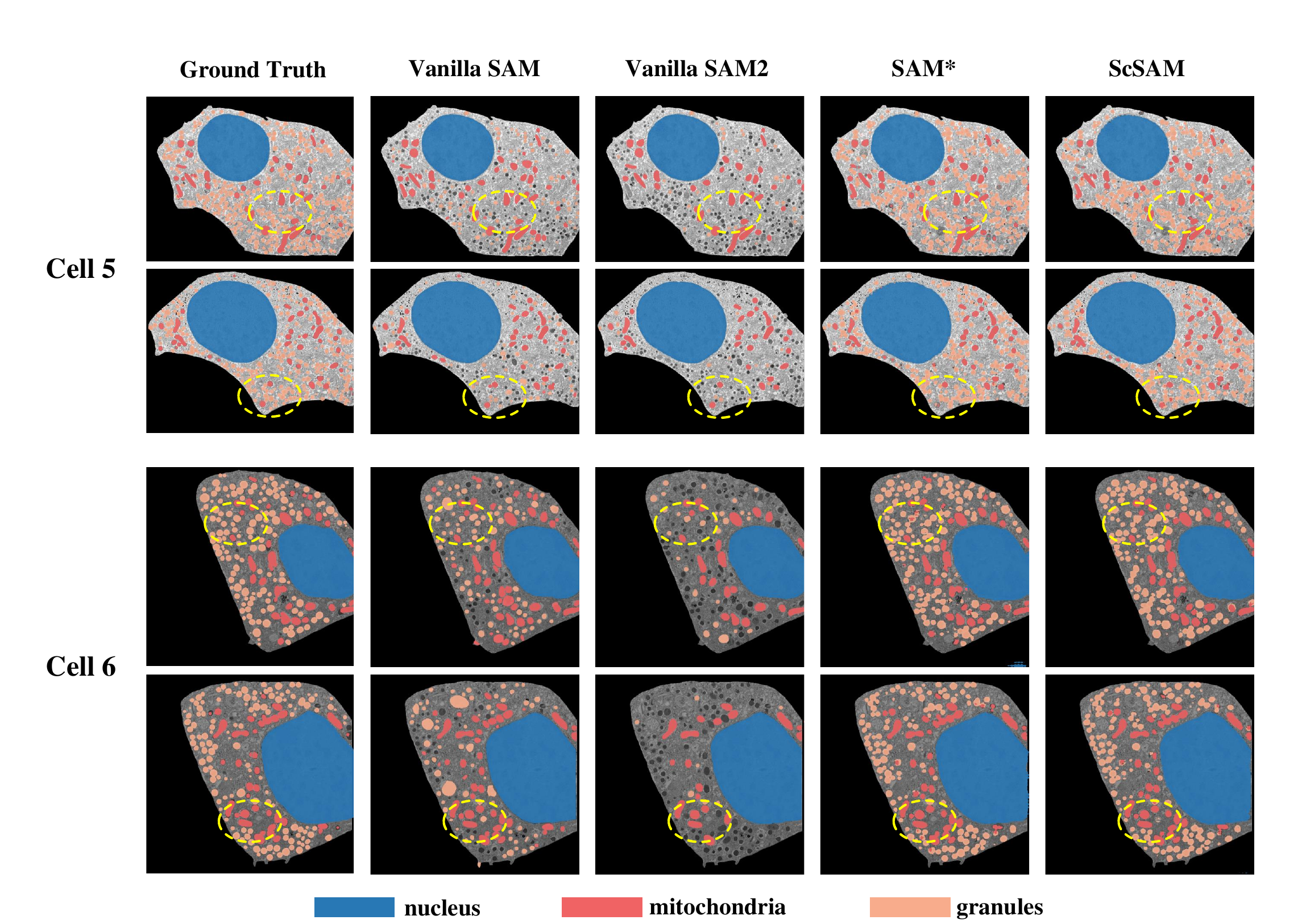}
	\caption{Visual comparison of prediction masks on low-glucose samples.}	
	\label{fig:result1} 
\end{figure*}

Fig.~\ref{fig:result} and Fig.~\ref{fig:result1} show the segmentation masks of each test cell, where two slices with a significant distance are selected for each cell to demonstrate the generalization and robustness of ScSAM.
Specifically, the substantial morphological and distributional variability of organelles in these slices can validate ScSAM's ability to distinguish and capture fine-grained details in complex EMIs.
Yellow circles are employed to highlight the segmentation mask of complex regions.

\subsection{Positive Class Similarity Visualization}
Fig.~\ref{fig:pc} compares the positive class similarity maps of islet cells across two datasets, visualizing the capture ability of ScSAM in edge details, internal textures, and complex shapes.
It can be observed that ScSAM enhances class-specific information representations when fusing embeddings, significantly reducing irrelevant features in the mitochondria and granules.


\begin{figure*}[t]
	\centering
	\includegraphics[width=\textwidth]{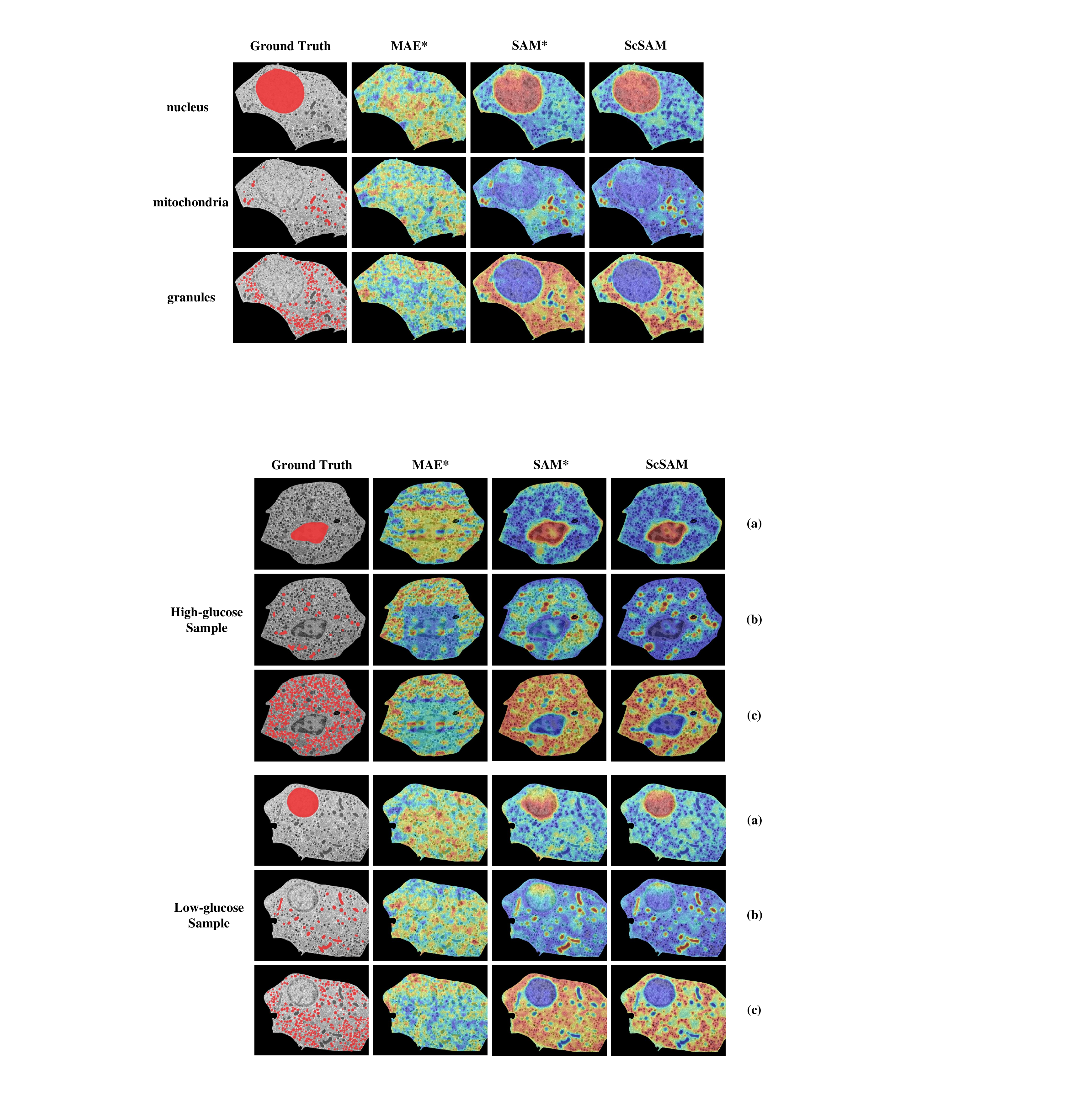}
	\caption{Visualization of positive class similarity maps for three categories: (a) nucleus; (b) mitochondria; (c) granules. We evaluate the cosine similarity between image embeddings and trained category prototypes on a pixel-by-pixel basis and overlay the similarity matrix over the original EMI for comparison.}	
	\label{fig:pc} 
\end{figure*}

\end{document}